\documentclass[11pt]{article}

\usepackage[preprint]{acl}

\usepackage{times}
\usepackage{latexsym}
\usepackage{booktabs}
\usepackage{amsmath}
\usepackage{amssymb}

\usepackage[T1]{fontenc}

\usepackage[utf8]{inputenc}

\usepackage{microtype}

\usepackage{inconsolata}

\usepackage{graphicx}

\title{No Tokenizer, Still a Gap: Cross-Lingual Fairness of Homoglyph and Invisible-Character Watermarking}

\author{
  Alexander Nemecek, Yuqiao Xu, Joseph Falkenburg, Erman Ayday \\
  Case Western Reserve University \\
  \texttt{\{ajn98, yxx914, jmf277, exa208\}@case.edu}
}

\begin{document}
\maketitle

\begin{abstract}
Homoglyph and invisible-character watermarks are edit-based watermarks. Instead of changing how a model generates text, they hide a signal in the text after it is written. They swap some characters for homoglyphs, which look the same, or very similar, but have different Unicode code points, or they insert characters that do not display, all without access to the model that wrote the text. Generation-time watermarks work less well for lower-resourced languages due to how tokenizers split their text. Because they work on individual characters, these watermarks never touch a tokenizer, so we ask whether they escape this gap. Across 40 language varieties in 24 scripts, we measure capacity (the hidden bits each character can carry), how much of it survives routine handling and targeted removal, and how much text a keyed detector needs. Capacity varies 3.3-fold and script accounts for 96\% of that variation. Within Latin script, resource level makes no detectable difference ($\rho=+0.095$, $p=0.78$), but across scripts capacity rises with it ($\rho=+0.54$, 95\% CI $[+0.14,+0.83]$), because lower-resourced languages tend to use scripts with few homoglyphs. Ordinary copy-paste, which strips invisible characters, widens the gap: a detector then needs about 2 code points of English or Persian but 93 of Lao. Removing the tokenizer does not remove the inequity, but instead only changes its source.
\end{abstract}

\section{Introduction}\label{sec:intro}
In April 2025, the education company Rumi reported that OpenAI's o3 and o4-mini models were leaving narrow no-break spaces (U+202F) in long answers~\cite{rumi2025watermarks}. These characters looked like ordinary spaces, survived pasting into Google Docs, and were taken for a hidden watermark. OpenAI said they were a side effect of training, but the case shows that people already treat disguised characters as evidence of where text came from.

\begin{figure}[htbp!]
    \centering
    \includegraphics[width=\columnwidth]{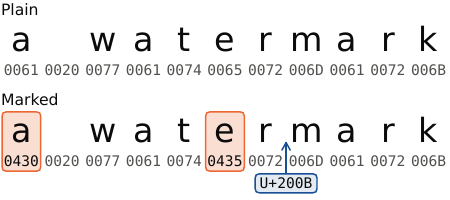}
    \caption{Two strings that render alike but differ in code points; the highlighted edits are two Cyrillic homoglyphs and a zero-width space.}
    \label{fig:homoglyph-example}
\end{figure}

Edit-based watermarks are built on such characters as they mark text after it is written, instead of while a model generates it. They rely on two kinds of character. The first is the homoglyph, a character that looks the same as another, but has a different Unicode code point. The letter \emph{a} as a Latin letter (U+0061) and a Cyrillic letter (U+0430), for example, look identical in most fonts, yet a computer stores them as different characters (Figure~\ref{fig:homoglyph-example}). The second is the invisible character, such as the zero-width space (U+200B), which takes up no room on screen.

A watermark swaps some characters for homoglyphs or inserts invisible ones, with a secret key deciding which edits to make~\cite{rizzo2016content, sato2023embarrassingly, por2012unispach, hellmeier2025innamark}. A detector holding the key can later check for that pattern, while the text still reads the same to a person. No access to the model that wrote the text is needed. Generation-time watermarks, by contrast, bias a model's sampling~\cite{kirchenbauer2023watermark, dathathri2024scalable, nemecek2026topic}, and post-hoc schemes that avoid model access still reword the text~\cite{chang2024postmark}.

Generation-time watermarks lose strength across languages as they do not survive translation~\cite{he2024crosslingual} and weaken in low- and medium-resource languages~\cite{mohamed2025multilingual}. The usual explanation is the tokenizer, which splits lower-resourced languages into more and smaller pieces and so shifts the statistics a detector relies on~\cite{petrov2023language, ahia2023all, rust2021good}. Edit-based watermarks never touch a tokenizer, which raises a simple question: without a tokenizer, does the gap remain? This has not been examined. Edit-based work reports capacity in aggregate or for a single script~\cite{shirali2008steganography, hellmeier2025innamark}, and work on watermark fairness covers statistical schemes only~\cite{nemecek2026gets}.

We cover 40 language varieties in 24 scripts. For each, we measure how many hidden bits each character can carry (its capacity), how much of that survives as the text is handled, how much text the detector needs, and how visible the edits are. Since the watermark is added after writing, it does not matter who wrote the text, so we use a human-translated parallel corpus where every language carries the same sentences, and differences reflect script and spelling, not content.

Marked text can change in two ways. Some changes are routine, such as copy-paste or the Unicode normalization many systems apply. They happen whether or not anyone is after the watermark, so if they erase more of it in some languages, that is a fairness problem. Other changes are deliberate. Security research uses the same characters to attack NLP systems and hide code~\cite{boucher2022bad, boucher2023trojan}, and Unicode specifies defenses that detect them or map them to a shared form~\cite{uts39}; we treat these defenses as the adversary. Combining that mapping with compatibility and whitespace folding erases the watermark in every language, so we report this floor rather than claim robustness.

Capacity turns out to be a property of the script. It varies more than threefold across languages and script accounts for almost all of that variation. Within Latin script, where enough languages exist for a direct test, resource level~\cite{joshi2020state} makes no detectable difference to capacity. Across scripts, however, lower-resourced languages tend to be written in scripts with few homoglyphs, so they receive less capacity. Ordinary copy-paste, which strips invisible characters, widens the gap: afterwards the same detector needs about 2 code points of English or Persian to find the watermark, but 93 of Lao. Removing the tokenizer does not remove the inequity; it moves its source into the Unicode standard. Our contributions are as follows:
\begin{enumerate}
    \item \textbf{A per-language framework} for the capacity, survival and visibility of edit-based watermarks across 40 language varieties in 24 scripts.
    \item \textbf{A transformation ladder} from routine handling to deliberate removal, with survival broken down by the kind of homoglyph or invisible character.
    \item \textbf{A keyed measurement instrument} showing that capacity predicts how much text a detector needs.
    \item \textbf{A fairness analysis} that treats scripts, not languages, as the unit of evidence, and shows that the disparity is organized by script yet still falls on lower-resourced languages.
\end{enumerate}

\section{Related Work}\label{sec:related_work}

\textbf{Generation-time watermarks across languages.} Generation-time watermarks bias which tokens a model samples~\cite{kirchenbauer2023watermark, kuditipudi2023robust, dathathri2024scalable, nemecek2026topic}, and they do not work equally well in every language. He et al.\ found that the signal does not survive translation and turned this into a removal attack~\cite{he2024crosslingual}. Mohamed and Gubri found that watermarks billed as multilingual fail in low- and medium-resource languages, and recover much of their strength when the text is routed through a better-supported language~\cite{mohamed2025multilingual}. The usual explanation is the tokenizer, which splits lower-resourced languages into more and smaller pieces~\cite{petrov2023language, ahia2023all, rust2021good} and so changes the per-token statistics a detector relies on. The watermarks we study never touch a tokenizer, so any gap they show must come from somewhere else.

\textbf{Edit-based watermarks.}
Edit-based watermarks change individual characters in finished text. Rizzo et al.\ watermark English text by homoglyph substitution~\cite{rizzo2016content}. Sato et al.'s EasyMark includes a variant for Chinese, Japanese and Korean text that uses variation selectors~\cite{sato2023embarrassingly}, so the field already knows the channel depends on script, but no comparison across scripts follows. Por et al.\ hide bits in the spaces between words~\cite{por2012unispach} and Hellmeier et al.\ benchmark ten whitespace methods on a million Wikipedia articles but report capacity as a single aggregate number~\cite{hellmeier2025innamark}. Shirali-Shahreza and Shirali-Shahreza hide data in Persian and Arabic text with zero-width joiner and non-joiner characters, within that one script~\cite{shirali2008steganography}. None of this work weights capacity by the characters a language actually uses, which is where the per-language gap appears.

\textbf{Homoglyphs as an attack.}
Security research treats the same characters as a threat. Boucher et al.\ showed that a few homoglyphs or invisible characters can break deployed NLP systems~\cite{boucher2022bad} and hide logic in source code~\cite{boucher2023trojan}. Unicode's own defense is older, UTS~\#39, written to catch spoofed identifiers such as domain names, lists confusables (Unicode's term for homoglyphs) and specifies mixed-script detection and skeleton matching, which maps homoglyphs to a shared form~\cite{uts39}. We use UTS~\#39 in both directions, utilizing its confusables list as our raw material and its detectors as the removal attacks in our ladder.

\textbf{Fairness in watermark evaluation.}
Nemecek et al.\ argue that watermark benchmarks should report results across languages and populations~\cite{nemecek2026gets, nemecek2026auditing}, but they cover statistical and frequency-domain watermarks and measure detection performance only. We study a character-level channel, measure its capacity directly, and trace its gap to Unicode's uneven supply of homoglyphs.

\section{Measurement Framework}\label{sec:framework}\label{sec:threat_model}

We treat an edit-based watermark as a channel and ask what it can carry in each language. An \textit{embedder} marks finished text with a secret key that decides which characters to swap and which gaps receive an invisible character; the edits aim to leave the text looking the same. A \textit{detector} holding the same key later checks a text. The detector also holds the original text, which makes it \textit{cover-informed}. It needs the original because a homoglyph does not reveal which character it replaced. Stated plainly, without the original, the detector cannot tell which substitutes a position had to choose from. A cover-informed detector is therefore an upper bound on one without it. This section defines the substitution sets, capacity and the transformations we apply. The sections that follow measure capacity and ask what organizes its differences, then measure how much survives, how much text a detector needs and how visible the edits are.

\textbf{Data.} We use the FLORES+ devtest split~\citep{nllb-24}, 1,012 sentences per language, for the 40 varieties listed in Table~\ref{tab:languages} (Appendix~\ref{app:fairness}). Character data are frozen at Unicode 17.0.0~\citep{unicodestandard, uts39} and CLDR 48.2.0.

\textbf{Two tracks and four families.}
The watermark is written on two tracks. The \textit{substitution} track replaces a character with a homoglyph in place; the \textit{insertion} track puts an invisible character into a gap between two characters. Prior systems use one track at a time (\S\ref{sec:related_work}); we model both. The substitution set $S(c)$ of a character $c$ holds $c$ itself and every character that can replace it, drawn from separate Unicode tables so that we can tell where capacity comes from. Family~(i), \emph{cross-script}, holds homoglyphs from another script, such as the Cyrillic a (U+0430) for the Latin a. These come from the UTS~\#39 confusables list, where two characters count as confusable when they map to the same skeleton~\citep{uts39}. Family~(ii), \emph{compatibility}, holds characters that Unicode's compatibility decomposition folds to $c$, such as the fullwidth A (U+FF21) for A. Family~(iv), \emph{whitespace}, holds the other White\_Space characters when $c$ is itself whitespace, such as the narrow no-break space (U+202F) from \S\ref{sec:intro}. Same-script confusables that are not compatibility variants form a \emph{residual} class, tracked separately. Family~(iii), the invisible characters, is the insertion alphabet $M$ rather than a substitution family; invisible characters never enter any $S(c)$, so the two tracks do not overlap. A substitute that fits several families is counted once, in the order iv, ii, i, residual. Homoglyphs of multi-character sequences cannot be swapped one for one and are excluded; Appendix~\ref{app:framework} gives the counts. Only family~(i) and the residual come from the hand-curated confusables list, which is known to be sparse for CJK and Indic scripts; families~(ii) and~(iv) come from systematic property tables.

\textbf{Admissibility.} Not every homoglyph suits every language, so three nested policies limit $S(c)$. Policy $\pi_{\mathrm{intra}}$ admits only substitutes in $c$'s own script; $\pi_{\mathrm{lang}}$ also admits any script the language normally uses, taken from its CLDR exemplar characters~\citep{cldr}; and $\pi_{\mathrm{none}}$ admits everything. Script-neutral characters (Script=Common or Inherited) pass under every policy. Because $\pi_{\mathrm{lang}}$ keeps the same-script clause, which matters for text embedded from another script such as Latin inside Japanese, the policies nest, $S_{\mathrm{intra}}(c)\subseteq S_{\mathrm{lang}}(c)\subseteq S_{\mathrm{none}}(c)$, and capacity can only grow from one to the next. We report $\pi_{\mathrm{none}}$ unless stated otherwise. The policies limit capacity; they do not guarantee stealth, which Appendix~\ref{app:stealth} tests.

\textbf{Capacity.} Capacity is the number of hidden bits one character of real text can carry, on average. A character with $k$ admissible forms (itself and $k-1$ substitutes) lets the key choose among $k$ options, which is $\log_2 k$ bits. Averaging over the language's character frequencies $p_L(c)$ gives the substitution capacity, in bits per character:
\begin{equation}
\kappa_{\mathrm{sub}}(L,\pi)=\sum_{c} p_L(c)\,\log_2 |S_\pi(c)|.
\end{equation}
Frequencies come from the raw corpus text, so spaces, punctuation, digits and characters borrowed from other scripts count at their true rate. A character with no substitute carries 0 bits, since $S(c)$ always contains $c$. Each gap between characters can hold nothing or one of the invisible characters in $M_L$, so the insertion track carries
\begin{equation}
\kappa_{\mathrm{ins}}(L)=g_L\,\log_2\bigl(|M_L|+1\bigr),
\end{equation}
where $g_L\approx 0.99$ is the number of gaps per character. $M_L$ holds ZERO WIDTH SPACE, ZERO WIDTH NON-JOINER, ZERO WIDTH JOINER and WORD JOINER, minus the two joiners in Arabic and in the Brahmic scripts that form conjuncts, where they change how letters join~\citep{w3c_alreq, unicodestandard}; $|M_L|$ is therefore 4 or 2. Total capacity is $\kappa=\kappa_{\mathrm{sub}}+\kappa_{\mathrm{ins}}$; for English at $\pi_{\mathrm{none}}$ it is 7.09 bits per character (4.78 from substitution and 2.30 from insertion, each rounded). Per-family capacities use each family's own set and do not add up to $\kappa_{\mathrm{sub}}$, because the logarithm of a union is not a sum of logarithms.

\begin{table}[htbp!]
\centering
\scriptsize
\begin{tabular}{@{}lll@{}}
\toprule
\textbf{Transformation} & \textbf{Operation} & \textbf{Removes} \\
\midrule
\multicolumn{3}{@{}l}{\textit{Routine}} \\
\texttt{identity}          & none                       & nothing \\
\texttt{nfc}               & canonical composition      & almost nothing \\
\texttt{copy\_paste}       & NFC, drop zero-width       & insertion track \\
\texttt{nfkc}              & compatibility composition  & compat.\ variants \\
\midrule
\multicolumn{3}{@{}l}{\textit{Deliberate (UTS \#39)}} \\
\texttt{mixed\_script\_hr} & detect-and-strip (HR)      & cross-script swaps \\
\texttt{mixed\_script\_ss} & detect-and-strip (SS)      & + embedded Latin \\
\texttt{skeleton}          & skeleton, keep ignorables  & listed confusables \\
\texttt{combined}          & canonical fold & both tracks \\
\bottomrule
\end{tabular}
\caption{Transformation ladder $\phi$. Routine transformations need no adversary; deliberate transformations use Unicode's own defenses.}
\label{tab:ladder}
\end{table}

\textbf{Transformations.} We also measure how much capacity survives the transformations in Table~\ref{tab:ladder}. Routine ones need no adversary: Unicode normalization (NFC, NFKC) and a lossy copy-paste, which we model as NFC plus the removal of zero-width characters. Some tools keep zero-width characters, so this level is the pessimistic case for the insertion track. Deliberate ones are Unicode's own defenses~\citep{uts39}, used by someone who wants the watermark gone while keeping the text usable: mixed-script detection, applied as detect-and-strip to each token (a run of non-whitespace characters), and the skeleton mapping, which folds homoglyphs into a shared form. The full UTS~\#39 skeleton also deletes default-ignorable characters, which include all four insertion characters; we keep the two steps apart so that each rung mainly hits one track, and the ladder ends with the full skeleton plus compatibility and whitespace folding, removing both tracks.

\begin{figure}[t!]
\centering
\includegraphics[width=\columnwidth]{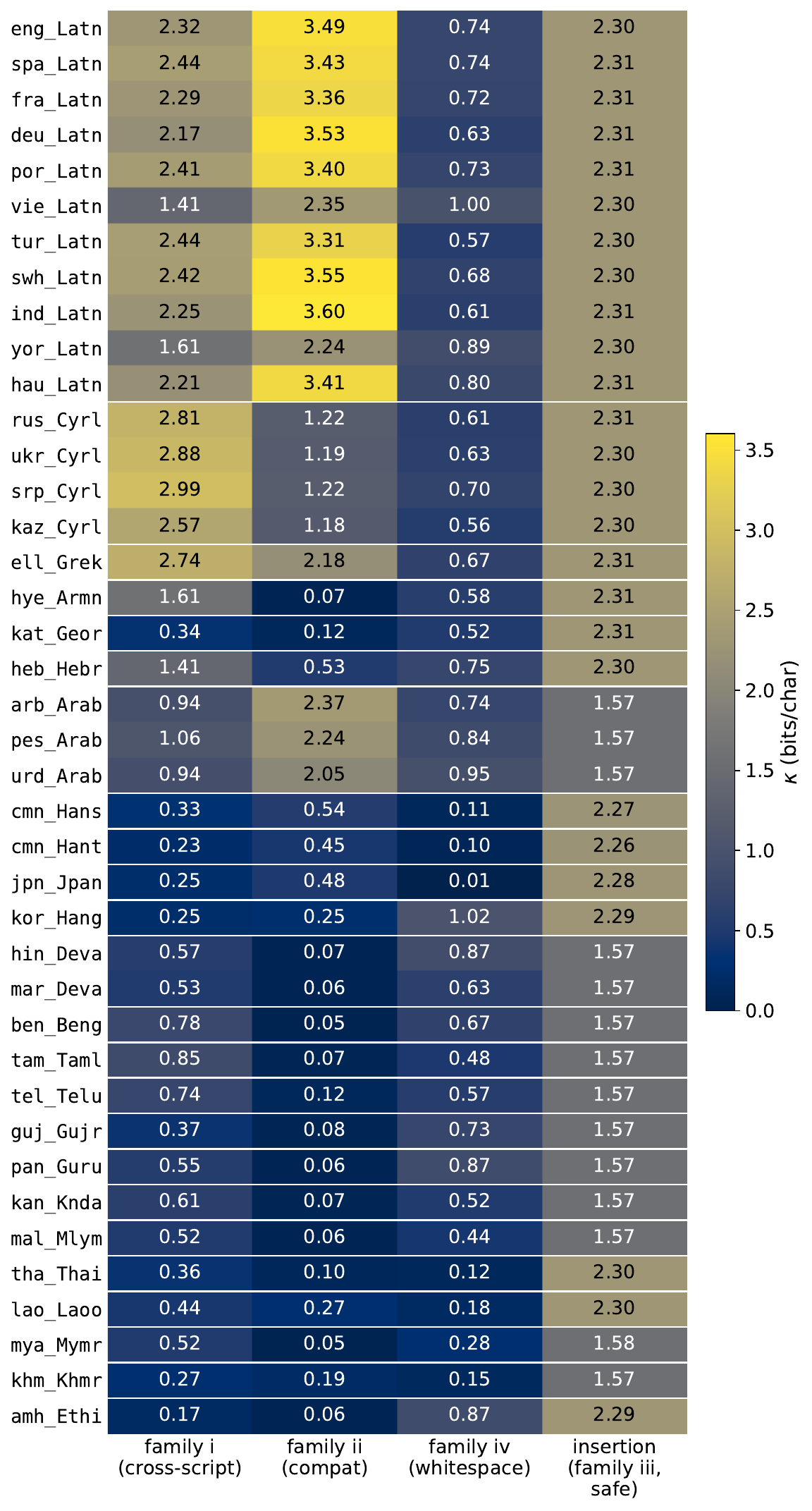}
\caption{Untransformed capacity under $\pi_{\mathrm{none}}$, in bits per character, by language (rows, grouped by script) and family (columns); the insertion column uses each language's safe invisible characters. Family values are computed separately and do not add up to the total.}\label{fig:capacity}
\end{figure}

\section{Capacity}\label{sec:capacity}

Figure~\ref{fig:capacity} shows capacity by family for every language before any transformation, and Table~\ref{tab:capacity-full} (Appendix~\ref{app:framework}) gives the full numbers by language, policy and track. Total capacity runs from 7.09 bits per character for Swahili and English down to 2.14 for Khmer, a 3.31-fold range (Gini coefficient 0.21). The range follows script lines since the mean is 6.81 bits per character for the 11 Latin-script languages, 3.21 for Chinese, Japanese and Korean, and 2.93 for the nine South Asian languages written in Brahmic scripts. These differences are not sampling noise, the bootstrap 95\% interval on English total capacity is $[7.082, 7.089]$ (Appendix~\ref{app:framework}).

The gap comes from the substitution track. Insertion depends only on how many invisible characters a language can safely use, not on its letters which are about 2.30 bits per character, or 1.57 in Arabic and in the scripts that form conjuncts, which must leave out the two joiners (\S\ref{sec:framework}). Substitution, by contrast, runs from under one bit per character to almost five (4.78 for English), because it depends on how many homoglyphs a language's frequent characters have.

Within substitution, compatibility variants (family~ii) do most of the separating. They are worth more than three bits per character for most Latin-script languages, whose letters and digits have many fullwidth and mathematical forms, and almost nothing for the Brahmic scripts. Cross-script homoglyphs (family~i) appear in every script but carry far less outside Latin, Cyrillic and Greek, three alphabets that share many letter shapes. Most conjunct-forming scripts are short on both tracks, which is why Khmer (2.14) and Myanmar (2.40) end lowest. Arabic, Persian and Urdu pay the same insertion penalty but still reach 5.5 to 5.8 bits per character, because Unicode encodes many Arabic presentation forms as compatibility variants.

Chinese, Japanese and Korean follow a different pattern. Their native characters have almost no homoglyphs, and characters outside their native character sets, such as Latin letters, digits and punctuation, carry 49 to 65\% of the substitution capacity of Chinese and Japanese and all of Korean's, against 18\% for English.

These capacities are high because $\pi_{\mathrm{none}}$ counts every substitute, including many that a cautious embedder would avoid. Dropping the cross-script substitutes that a mixed-script detector would flag ($\pi_{\mathrm{intra}}$), such as a Cyrillic a in English text, leaves Latin-script languages 93 to 95\% of their substitution capacity but the Brahmic and South-East Asian ones only 54 to 81\%. Keeping only substitutes that UTS~\#39 marks as \emph{Allowed} in identifiers leaves English 1.42 of its 4.78 substitution bits.

Curation explains part of the gap, as \S\ref{sec:framework} notes, family~(i) and the residual come from a hand-curated list that is sparse for CJK and Indic scripts. But family~(ii), which comes from a systematic table, splits by script just as sharply. Unicode added compatibility variants largely to stay compatible with older character sets and to support mathematical styling, needs that covered Latin, Greek, Arabic and East Asian forms but hardly any Brahmic letters. The gap is therefore a property of the Unicode standard, not of our method.

\section{Fairness}\label{sec:fairness}

Our unit of evidence is the script, not the language. The 11 Latin-script languages are close copies of one another, so counting them as 11 independent observations would overstate the evidence; every statement about scripts therefore rests on 24 script groups. We measure how much of the variation in capacity lies between scripts with the intraclass correlation ICC(1)~\citep{shrout1979intraclass}, which is 1 when languages that share a script have identical capacity and near 0 when script tells us nothing (Appendix~\ref{app:fairness}). Confidence intervals resample whole scripts rather than languages~\citep{field2007bootstrapping}.

\begin{figure}[h!]
\centering
\includegraphics[width=\columnwidth]{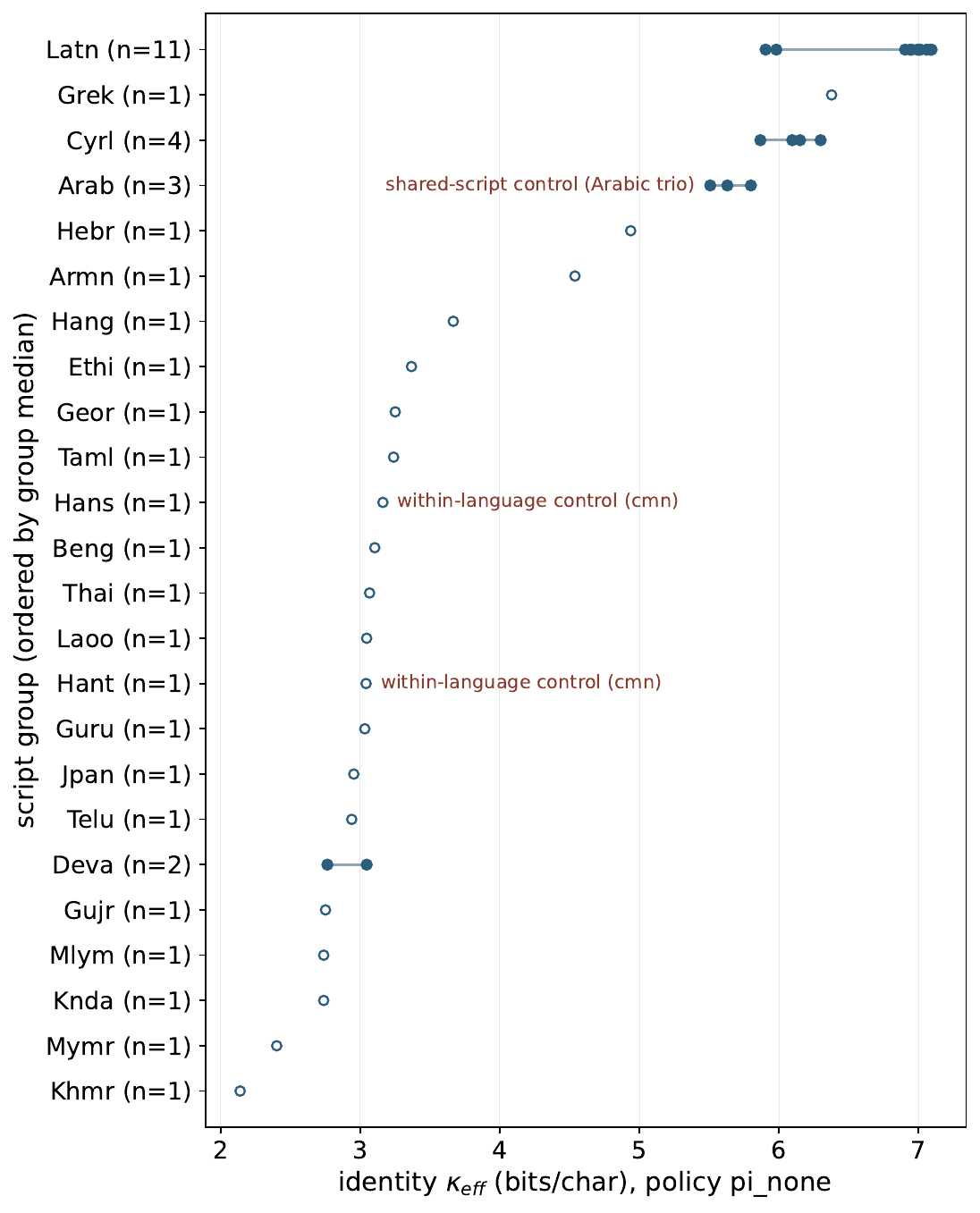}
\caption{Untransformed capacity under $\pi_{\mathrm{none}}$ by script group, ordered by group median. Bars span the languages of the four scripts that have several; open markers are scripts with one language.}\label{fig:byscript}
\end{figure}

Script explains almost all of the variation. The ICC is 0.963, so the variance between scripts is about 25 times the variance within them, and none of 10,000 shuffles of languages across scripts, with group sizes fixed, reached it ($p<10^{-4}$; Figure~\ref{fig:byscript}). The ICC stays between 0.954 and 0.978 under each transformation in Figure~\ref{fig:effective} and between 0.963 and 0.967 across the three policies so the result does not rely on one setting.

The four scripts with several languages show how little room is left inside a script. Arabic, Persian and Urdu sit in three different resource classes, yet their capacities span only 0.29 bits per character. The 11 Latin-script languages span 1.19 bits, against about 5 bits between scripts, and most of that spread comes from Vietnamese and Yoruba, which write many accented letters with few substitutes. Changing only the characters, by contrast, moves capacity, for example, Mandarin in Simplified and in Traditional characters, one language in one resource class, differs by 1.04-fold, because insertion is the same for both, and by 1.24-fold once copy-paste removes insertion.

To ask whether lower-resourced languages receive less capacity, we use the resource classes of~\citet{joshi2020state}, from 0 (least resourced) to 5, listed in Table~\ref{tab:languages} (Appendix~\ref{app:fairness}). Only Latin script has enough languages for a direct test inside one script. Before looking at any class values, we required that test to cover at least three classes spanning at least two levels; the 11 Latin-script languages cover four classes spanning three. The rank correlation between capacity and class is $\rho=+0.095$ (permutation $p=0.78$), and the class means of 6.66, 7.01, 6.63 and 7.01 bits per character for classes 2 to 5 show no trend.

\begin{figure}[h!]
\centering
\includegraphics[width=\columnwidth]{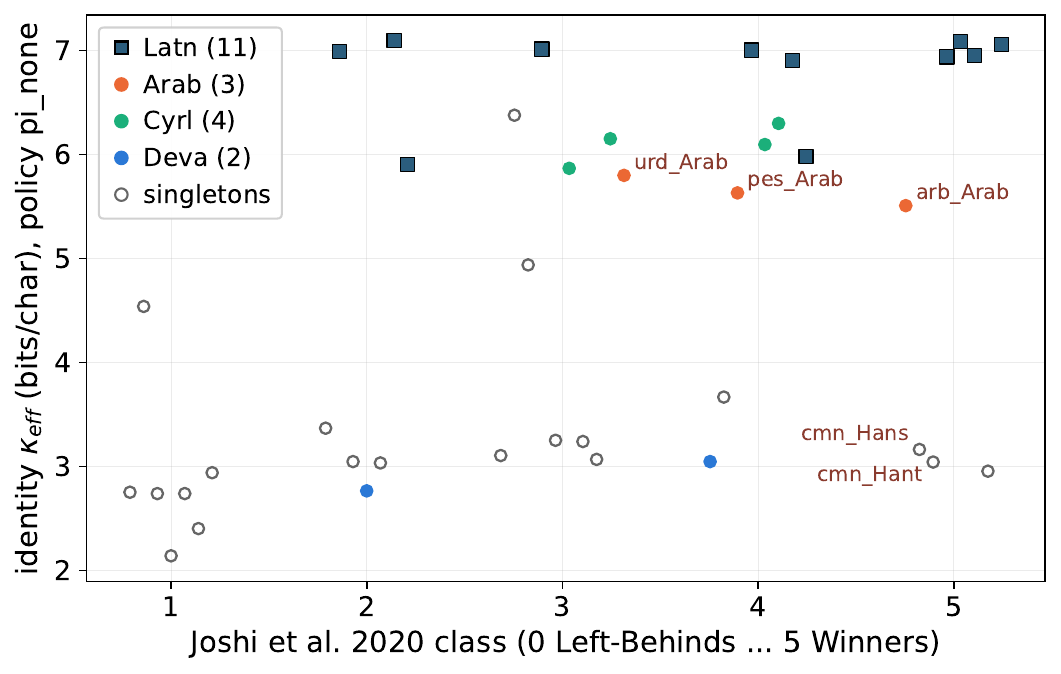}
\caption{Untransformed capacity under $\pi_{\mathrm{none}}$ against the resource class of~\citet{joshi2020state}. Horizontal offsets within a class only separate the points.}\label{fig:resource}
\end{figure}

Across scripts, a script's median capacity rises with its median resource class ($\rho=+0.54$, 95\% interval $[+0.14,+0.83]$ over 24 scripts; Figure~\ref{fig:resource}). Low-capacity scripts, such as the Brahmic, South-East Asian and Ethiopic ones, are used mostly by languages in classes 1 to 3, while Latin, Cyrillic and Arabic hold most of the languages in classes 4 and 5. Chinese, Japanese and Korean, in classes 4 and 5 but with little capacity, are the main exception, which keeps the correlation moderate. Table~\ref{tab:fairness} collects these results.

\begin{table}[h!]
\centering\scriptsize
\setlength{\tabcolsep}{4pt}
\begin{tabular}{@{}lll@{}}
\toprule
Quantity & Value & Evidence base \\
\midrule
ICC, untransformed & 0.963, $p<10^{-4}$ & 24 scripts \\
ICC, per transformation & 0.954--0.978 & 24 scripts \\
Mandarin, two orthographies & $1.04\times$ (copy-paste $1.24\times$) & 1 contrast \\
Within Latin, $\rho$ vs.\ class & $+0.095$, $p=0.78$ & 11 languages \\
Script level, $\rho$ vs.\ class & $+0.544$ $[+0.135,+0.826]$ & 24 scripts \\
Language level, $\rho$ vs.\ class & $+0.487$ $[+0.185,+0.740]$ & 40 languages \\
\bottomrule
\end{tabular}
\caption{Fairness results. $\rho$ is the rank correlation between capacity and resource class. Between-script rows count script groups as the evidence; intervals are 95\% bootstraps that resample whole scripts, and the language-level row clusters its 40 languages by script.}
\label{tab:fairness}
\end{table}

Taken together, script organizes the gap, and resource level lines up with script. At the script level the two cannot be separated, because each script contributes a single point. We therefore do not claim that resource level is irrelevant, only that its link to capacity appears between scripts and not within them. This answers the question from \S\ref{sec:intro} since without a tokenizer the gap remains, but its source is Unicode's uneven supply of homoglyphs rather than tokenization.

\section{Robustness}\label{sec:robustness}

\begin{figure}[t!]
\centering
\includegraphics[width=\columnwidth]{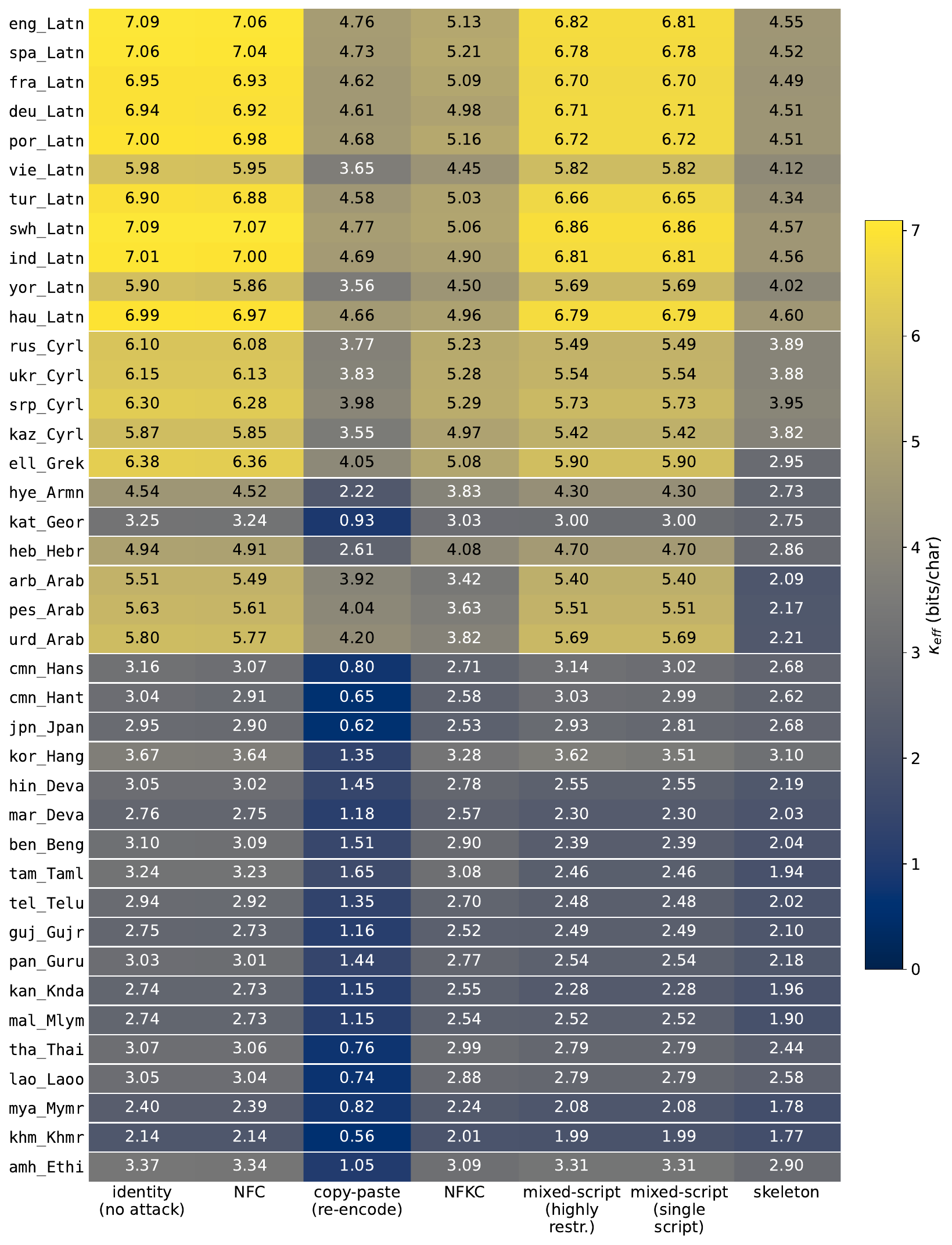}
\caption{Effective capacity under $\pi_{\mathrm{none}}$, in bits per character with both tracks summed, for each language and transformation. The first column is the untransformed total of Table~\ref{tab:capacity-full}; the skeleton column keeps invisible characters (Table~\ref{tab:ladder}).}\label{fig:effective}
\end{figure}

A transformation can turn two different substitutes into the same character and the key can then no longer tell them apart. Effective capacity therefore counts only the distinct results that survive. For a transformation $\phi$ from Table~\ref{tab:ladder}, applied character by character,
\begin{multline}
\kappa^{\mathrm{eff}}_{\mathrm{sub}}(L,\pi,\phi) \\
= \sum_c p_L(c)\log_2\bigl|\{\phi(c'):c'\in S_\pi(c)\}\bigr|,
\end{multline}
and likewise $\kappa^{\mathrm{eff}}_{\mathrm{ins}}=g_L\log_2|\{\phi(m):m\in M_L\}\cup\{\varnothing\}|$, where a deleted invisible character counts as no insertion. Survival $s=\kappa^{\mathrm{eff}}/\kappa$ is the share that remains. Figure~\ref{fig:effective} gives $\kappa^{\mathrm{eff}}$ for every language and transformation under $\pi_{\mathrm{none}}$, and Table~\ref{tab:rho} (Appendix~\ref{app:robustness}) gives $s$ for each track.

The two tracks fail under different transformations. Copy-paste deletes every inserted character, so the insertion track falls to zero in all 40 languages, while visible substitutes pass through. The skeleton mapping folds cross-script and residual homoglyphs into one shared form but keeps invisible characters (\S\ref{sec:framework}). On average, total capacity keeps 0.99 of its value under NFC, 0.93 under mixed-script detection, 0.83 under NFKC, 0.68 under skeleton mapping and 0.51 under copy-paste, but which one hurts most depends on the language. The combined attack of Table~\ref{tab:ladder} erases the mark in every language (\S\ref{sec:keyed}).

NFKC is usually assumed to neutralize homoglyphs, but it folds only compatibility variants~\citep{uax15}. It removes 96\% of compatibility pairs yet keeps 93\% of cross-script pairs, since a Cyrillic a is a separate letter, not a variant of the Latin one. Languages lose in proportion to how much of their capacity came from compatibility variants (Pearson $r=-0.84$): Chinese, Japanese and the Arabic-script languages keep 0.36 to 0.53 of their substitution capacity, the Latin-script languages 0.55 to 0.61, and the Brahmic and South-East Asian scripts, with few such variants to lose, 0.77 to 0.91.

Mixed-script detection is a defense aimed at homoglyphs which might be expected to hurt homoglyph-rich Latin text most, yet at the Highly Restrictive level it keeps 0.94 to 0.98 of the substitution capacity of Latin, Arabic and CJK text but only about 0.65 for the Brahmic and South-East Asian scripts. Brahmic homoglyphs are mostly twins from sibling scripts, such as Bengali and Tirhuta, and swapping in a sibling-script letter is exactly the mixing the detector strips.

The disparity across all 40 languages can be compared even though no single robustness ranking fits all. NFKC and skeleton mapping narrow the 3.31-fold range to 2.63 and 2.59, because they remove what made Latin and Arabic rich; mixed-script detection widens it slightly, to 3.45. Copy-paste widens it most, to 8.49 (Gini 0.34), because insertion is most of what substitution-poor languages have and removing it costs Khmer 74\% of its capacity but Swahili only 33\%.

\begin{figure*}[ht!]
\centering
\includegraphics[width=\textwidth]{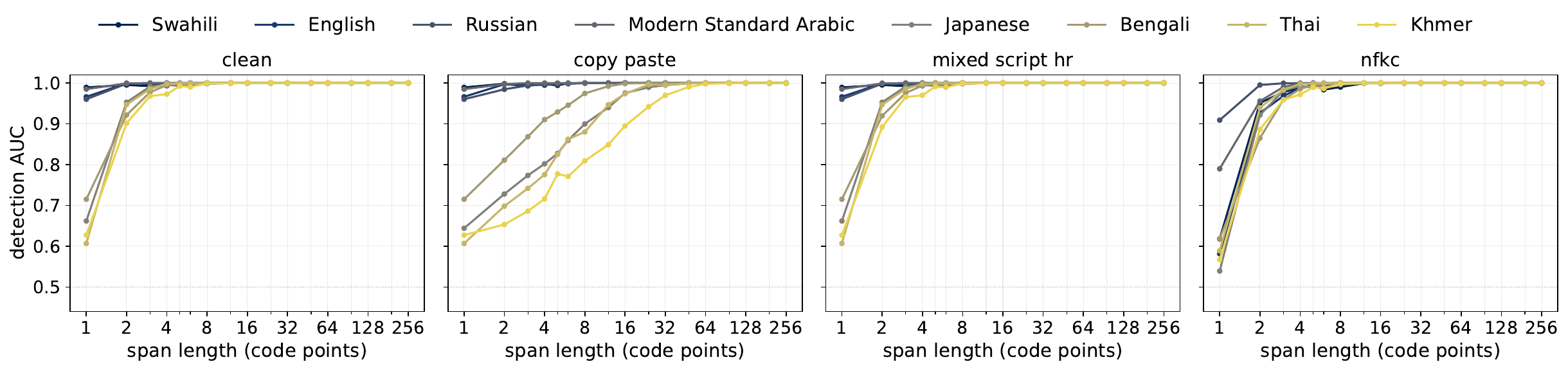}
\caption{Detection AUC against span length for eight languages across the capacity range, under four attacks.}
\label{fig:auc}
\end{figure*}

\begin{figure*}[ht!]
\centering
\includegraphics[width=\textwidth]{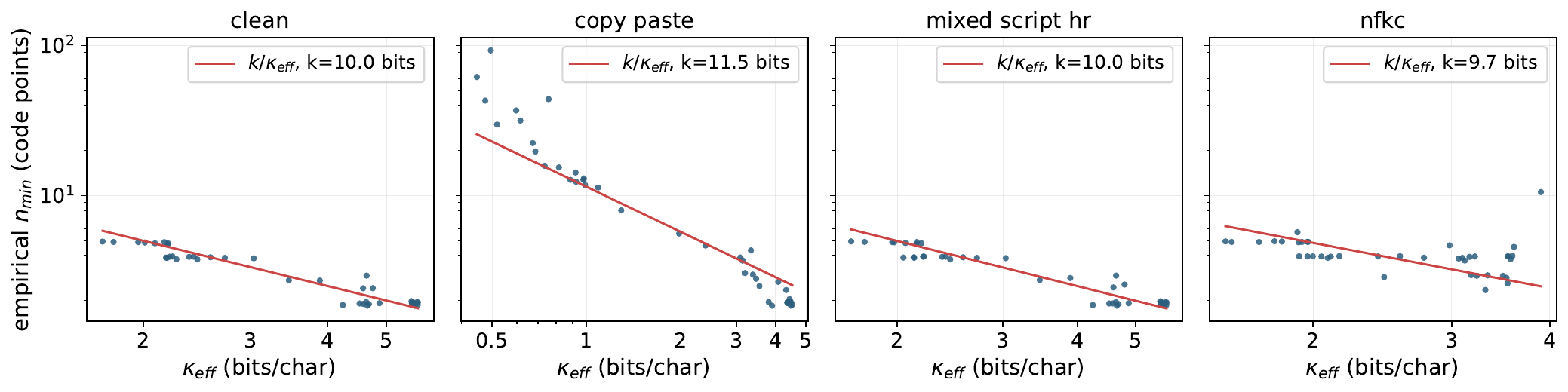}
\caption{Measured $n_{\min}$ (TPR 95\% at FPR 1\%) against $\kappa^{\mathrm{eff}}$ at that length, one point per language. Lines are $k/\kappa^{\mathrm{eff}}$ with $k$ fitted once per attack (policy $\pi_{\mathrm{lang}}$).}
\label{fig:nmin}
\end{figure*}

\textbf{Stealth.} Appendix~\ref{app:stealth} checks three ways a substitute can be noticed: a mixed-script detector, Unicode's identifier status, and the rendered glyph, which depends on the font (Figure~\ref{fig:fontdiv}). As with capacity, the cost of staying hidden follows script, for instance, Tamil pays through Unicode's script assignments, Bengali through a sibling script that readers' fonts lack, and Arabic through how one shaping engine treats invisible characters.

\section{Keyed Validation}\label{sec:keyed}

Bits per character are an accounting quantity until a detector cashes them. We therefore build a keyed instrument on the measured channel and ask how much text a detector needs. The instrument per-slot alphabets are exactly the admissible sets of \S\ref{sec:framework}. A PRNG seeded by the key draws one symbol for each substitution slot and each insertion gap. The detector holds the key and the cover (\S\ref{sec:framework}), rebuilds the schedule from the received text, and counts matching slots, much as generation-time detectors count green-list tokens~\citep{kirchenbauer2023watermark}. It thresholds the count's z-score at the 99th percentile of null trials; Appendix~\ref{app:keyed} describes the canonical keying and the test.

\textbf{Sweep.}
For each language we score spans of 1 to 256 code points under clean text, copy-paste, the mixed-script detector and NFKC, at policy $\pi_{\mathrm{lang}}$; a span with no usable slot counts as zero evidence rather than being skipped. $n_{\min}$ is the shortest span reaching a true-positive rate of 95\% at a false-positive rate of 1\% (Figures~\ref{fig:auc} and~\ref{fig:nmin}; Table~\ref{tab:nmin}). Here $\kappa^{\mathrm{eff}}$ is evaluated at $\pi_{\mathrm{lang}}$ and at the gap rate of a span of length $n_{\min}$, because that is what the detector sees; for English it is 5.63 bits per character rather than the 7.09 of \S\ref{sec:capacity}.

\textbf{Capacity predicts detection length.}
On clean text $n_{\min}$ runs from 1.85 code points (Persian) to 4.95 (Khmer). A single constant reproduces this as $n_{\min}\approx k/\kappa^{\mathrm{eff}}$ with $k = 10.02$ bits, the median of $n_{\min}\kappa^{\mathrm{eff}}$ on clean text, and the measured-to-predicted ratio runs from 0.79 to 1.36. The rank correlation between $n_{\min}$ and $1/\kappa^{\mathrm{eff}}$ is 0.877 clean, 0.969 under copy-paste and 0.863 under the mixed-script detector; under NFKC it is 0.552, with Turkish an outlier at 10.57 code points (interval 2.99 to 12.82). Under copy-paste the $n_{\min}$ range opens to 50$\times$, English and Persian still need about 2 code points, but Lao needs 92.96.

\textbf{Floor.}
The combined attack of Table~\ref{tab:ladder} is a true floor with detection at chance for all 40 languages (largest AUC 0.589).

\section{Conclusion}\label{sec:conclusion}
Homoglyph and invisible-character watermarks need no tokenizer, yet they do not escape the resource gap of generation-time watermarks. Across 40 language varieties in 24 scripts, script accounts for 96\% of the variation in capacity and for over 95\% under every attack but the floor. Resource level makes no detectable difference within Latin script, but lower-resourced languages tend to use scripts for which Unicode lists few homoglyphs, hence less capacity. Capacity predicts the text a detector needs for a copy-paste that strips invisible characters widens the gap from 3.3-fold to 8.5-fold; the detector then needs about 2 code points of English or Persian but 93 of Lao. An adversary can remove the watermark in any language; our concern is unequal protection under routine handling. Narrowing the gap would require more complete confusables data for under-served scripts, or a post-hoc channel independent of script-specific inventories.

\section*{Limitations}\label{sec:limitations}

\textbf{Design.}
Twenty of the 24 script groups hold a single language, so within-script variance is estimated from four groups and 16 residual degrees of freedom. The ICC is high partly because most groups cannot contribute within-group variance at all; the permutation test preserves group sizes and accounts for this, but the limitation remains. At the script level, each script contributes one median, so script identity and resource class cannot be separated there, and the positive script-level association cannot be attributed to either. The only within-script test (Latin, $n=11$, four classes) meets the coverage rule we fixed before examining any class values (\S\ref{sec:fairness}), but at that size a null result can only bound an effect, not rule one out.

\textbf{Resource taxonomy.}
We use one published classification~\citep{joshi2020state}, matched to our languages by name. Seven of the 40 matches are judgment calls, and one exact match, Serbian, also has a plausible lower-class alternative (Appendix~\ref{app:fairness}). A different taxonomy or matching could change the resource classes, but not the grouping by script.

\textbf{Corpus.}
We use one parallel corpus in one register. Parallel text makes the languages directly comparable but limits how far the results generalize. As a check, recomputing on the FLORES+ dev split changes substitution capacity by 0.012 bits per character on average and by at most 0.060.

\textbf{Snapshot.}
All capacities are computed from Unicode 17.0.0, CLDR 48.2.0 and FLORES+ 4.5. The confusables list grows between versions, so every capacity here is specific to these versions. Python's normalization library uses older Unicode data (version 14.0); 158 of the 8,275 code points in our substitution sets are newer than that, including Cyrillic modifier letters from Unicode 15 that substitute for frequent Cyrillic letters. The library's NFKC cannot fold these newer variants, which is why it removes only 96\% of compatibility pairs (\S\ref{sec:robustness}) and leaves the four Cyrillic-script languages more NFKC capacity than Unicode 17.0 normalization would. We keep the library for the NFKC attack, so that its results match what the library does in practice, and use the Unicode 17.0.0 decompositions for the combined floor attack.

\textbf{Renderer dependence.}
Every perceptual number depends on the fonts used. When a font stack draws a substitute with a different font than the original character, the measured distance reflects the stack's fixed fallback order rather than any one device. Whether an invisible character stays invisible depends on the shaping engine, most sharply for Arabic (1.57 bits per character under OpenType and Apple's CoreText, 0 when HarfBuzz shapes Apple's AAT fonts), and 178 of the 320 combinations of language, engine and character are not measured: 32 are set by rule and 146 are assumed safe without a test. The 0.02 threshold for eye-invisible substitutes is a choice, and results are stable only over a narrow range around it, so Table~\ref{tab:stealth} also reports bit-identical substitutes, which need no threshold. No human perceptual study was run; all perceptual claims are computational.

\textbf{Channel scope and detector.}
We exclude the 1,207 code points that are confusable only with multi-character sequences, since they cannot be swapped one for one, and we leave variation selectors out of the capacity totals. The token-level detector is the weakest reasonable one; the per-pair bound differs from it by at most 0.03 bits per character. The keyed detector is cover-informed, so it is an upper bound on a detector without the original text; within that setting, its threshold is set empirically from null trials, and Appendix~\ref{app:keyed} reports its false-positive rate on fresh keys. In 22 languages, the text already contains some of the invisible characters we insert, so some insertions cannot be told apart from native ones; this matters only for Khmer.

\section*{Ethical Considerations}
Homoglyph and zero-width text manipulation is publicly known and documented in the security literature and in Unicode's own standards~\citep{boucher2022bad, boucher2023trojan, uts39}; our substitution sets are derived from published Unicode data files. This paper contributes a measurement framework and a validation instrument, not a watermarking product or an attack tool: the keyed instrument uses only the character sets defined in \S\ref{sec:framework}, so it carries exactly the capacity we measure and nothing more. The harm we document is unequal protection rather than a new exploit. Text in low-capacity scripts is harder to mark and harder to verify, so provenance tools built on this channel silently protect some language communities less than others; documenting that gap is the point of this paper. The transformation ladder and our detector models are directly useful to those building normalization and detection defenses, and the per-language numbers show deployers where coverage is thin. We use only public, licensed data: the FLORES+ corpus (CC BY-SA 4.0) and public Unicode and CLDR files. No personal data, user content or human subjects are involved; the objects of study are scripts and code points, not communities. At the corpus provider's request, we do not redistribute raw corpus sentences; every released artifact is a derived statistic that can be rebuilt exactly from fixed data versions, and we document how the resource classes were matched. Nothing here recommends deploying this channel or certifies any configuration as safe.



\bibliography{custom}

\appendix

\begin{table}[t!]
\centering\footnotesize
\setlength{\tabcolsep}{2.5pt}
\begin{tabular}{@{}lllc@{}}
\toprule
Code & Language & Script & Class \\
\midrule
\texttt{eng\_Latn} & English & Latin & 5 \\
\texttt{spa\_Latn} & Spanish (Lat.\ Am.) & Latin & 5\rlap{$^{\dagger}$} \\
\texttt{fra\_Latn} & French & Latin & 5 \\
\texttt{deu\_Latn} & German & Latin & 5 \\
\texttt{por\_Latn} & Portuguese (Brazilian) & Latin & 4\rlap{$^{\dagger}$} \\
\texttt{vie\_Latn} & Vietnamese & Latin & 4 \\
\texttt{tur\_Latn} & Turkish & Latin & 4 \\
\texttt{swh\_Latn} & Swahili & Latin & 2 \\
\texttt{ind\_Latn} & Indonesian & Latin & 3 \\
\texttt{yor\_Latn} & Yoruba & Latin & 2 \\
\texttt{hau\_Latn} & Hausa & Latin & 2 \\
\addlinespace[2pt]
\texttt{rus\_Cyrl} & Russian & Cyrillic & 4 \\
\texttt{ukr\_Cyrl} & Ukrainian & Cyrillic & 3 \\
\texttt{srp\_Cyrl} & Serbian & Cyrillic & 4 \\
\texttt{kaz\_Cyrl} & Kazakh & Cyrillic & 3 \\
\addlinespace[2pt]
\texttt{ell\_Grek} & Greek & Greek & 3 \\
\addlinespace[2pt]
\texttt{hye\_Armn} & Armenian & Armenian & 1 \\
\addlinespace[2pt]
\texttt{kat\_Geor} & Georgian & Georgian & 3 \\
\addlinespace[2pt]
\texttt{heb\_Hebr} & Hebrew & Hebrew & 3 \\
\addlinespace[2pt]
\texttt{arb\_Arab} & Modern Standard Arabic & Arabic & 5\rlap{$^{\dagger}$} \\
\texttt{pes\_Arab} & Western Persian & Arabic & 4\rlap{$^{\dagger}$} \\
\texttt{urd\_Arab} & Urdu & Arabic & 3 \\
\addlinespace[2pt]
\texttt{cmn\_Hans} & Mandarin Chinese & Han (Simpl.) & 5\rlap{$^{\dagger}$} \\
\addlinespace[2pt]
\texttt{cmn\_Hant} & Mandarin Chinese & Han (Trad.) & 5\rlap{$^{\dagger}$} \\
\addlinespace[2pt]
\texttt{jpn\_Jpan} & Japanese & Han and kana & 5 \\
\addlinespace[2pt]
\texttt{kor\_Hang} & Korean & Hangul & 4 \\
\addlinespace[2pt]
\texttt{hin\_Deva} & Hindi & Devanagari & 4 \\
\texttt{mar\_Deva} & Marathi & Devanagari & 2 \\
\addlinespace[2pt]
\texttt{ben\_Beng} & Bengali & Bengali & 3 \\
\addlinespace[2pt]
\texttt{tam\_Taml} & Tamil & Tamil & 3 \\
\addlinespace[2pt]
\texttt{tel\_Telu} & Telugu & Telugu & 1 \\
\addlinespace[2pt]
\texttt{guj\_Gujr} & Gujarati & Gujarati & 1 \\
\addlinespace[2pt]
\texttt{pan\_Guru} & Eastern Panjabi & Gurmukhi & 2\rlap{$^{\dagger}$} \\
\addlinespace[2pt]
\texttt{kan\_Knda} & Kannada & Kannada & 1 \\
\addlinespace[2pt]
\texttt{mal\_Mlym} & Malayalam & Malayalam & 1 \\
\addlinespace[2pt]
\texttt{tha\_Thai} & Thai & Thai & 3 \\
\addlinespace[2pt]
\texttt{lao\_Laoo} & Lao & Lao & 2 \\
\addlinespace[2pt]
\texttt{mya\_Mymr} & Burmese & Myanmar & 1 \\
\addlinespace[2pt]
\texttt{khm\_Khmr} & Khmer & Khmer & 1 \\
\addlinespace[2pt]
\texttt{amh\_Ethi} & Amharic & Ethiopic & 2 \\
\bottomrule
\end{tabular}
\caption{The 40 language varieties. Class is the resource class of~\citet{joshi2020state}, from 0 (least resourced) to 5 (most); $\dagger$ marks a class assigned by a judgement call rather than an exact name match.}
\label{tab:languages}
\end{table}

\section{Languages and resource classes}\label{app:fairness}
Table~\ref{tab:languages} lists the 40 varieties of \S\ref{sec:framework}. The code is the FLORES+ identifier; it labels the rows of Figures~\ref{fig:capacity} and~\ref{fig:effective} and of every table below, and its suffix is the script, which gives the 24 script groups of \S\ref{sec:fairness}. The last column is the resource class used there \citep{joshi2020state}. Classes are joined by language name: 33 are exact matches and seven, marked $\dagger$, are judgement calls. Two drop a regional qualifier (Spanish, Portuguese); three match across a naming difference (Modern Standard Arabic to Arabic, Western Persian to Persian, Eastern Panjabi to Eastern Punjabi); and both Mandarin varieties take the single Mandarin row, so that pair shares a class by construction. Serbian takes its exact row (class~4), although Serbo-Croatian rows of lower class also exist.

\textbf{Fairness analysis.}
We use the one-way random-effects intraclass correlation for unbalanced groups~\citep{shrout1979intraclass},
\begin{equation*}
\mathrm{ICC}(1)=\frac{\mathrm{MSB}-\mathrm{MSW}}{\mathrm{MSB}+(k_0-1)\,\mathrm{MSW}},
\end{equation*}
where MSB and MSW are the between- and within-script mean squares and $k_0$ is the adjusted mean group size. Permutation tests shuffle the language-to-script labels while keeping group sizes; seeds are listed in Appendix~\ref{app:repro}. The variance between scripts is 3.31 and the variance within them 0.13, a ratio of about 25. The ICC stays high under every transformation because each transformation removes particular kinds of homoglyph, and languages that share a script share those kinds, so their capacities rise and fall together.

\textbf{Descriptive covariates.}
Three script-level quantities, read from the frozen Unicode tables and never from the corpus or the substitution sets, describe the scripts further. We exclude any quantity built from the substitution sets, since explaining capacity with the same confusables data that defines it would be circular; these covariates therefore describe the picture rather than confirm the mechanism. Against script-median capacity over 24 scripts, script size (code points assigned) has Spearman $+0.35$ $[+0.01,+0.63]$; the five bicameral scripts, which have upper and lower case, exceed the rest by 3.08 bits per character in median $[+0.21,+3.95]$; and combining-mark density against substitution capacity is $-0.06$ $[-0.48,+0.39]$. The three are correlated with one another (script size against mark density, $-0.56$), so they should be read as one picture rather than three findings.

\begin{table}[h!]
\centering\footnotesize
\setlength{\tabcolsep}{4pt}
\begin{tabular}{@{}lccccc@{}}
\toprule
 & \multicolumn{3}{c}{$\kappa_{\mathrm{sub}}$} & & \\
\cmidrule(lr){2-4}
Code & $\pi_{\mathrm{intra}}$ & $\pi_{\mathrm{lang}}$ & $\pi_{\mathrm{none}}$ & $\kappa_{\mathrm{ins}}$ & $\kappa$ \\
\midrule
\texttt{eng\_Latn} & 4.50 & 4.50 & 4.78 & 2.30 & 7.09 \\
\texttt{spa\_Latn} & 4.46 & 4.46 & 4.75 & 2.31 & 7.06 \\
\texttt{fra\_Latn} & 4.38 & 4.38 & 4.64 & 2.31 & 6.95 \\
\texttt{deu\_Latn} & 4.39 & 4.39 & 4.63 & 2.31 & 6.94 \\
\texttt{por\_Latn} & 4.39 & 4.39 & 4.70 & 2.31 & 7.00 \\
\texttt{vie\_Latn} & 3.50 & 3.50 & 3.68 & 2.31 & 5.98 \\
\texttt{tur\_Latn} & 4.34 & 4.34 & 4.60 & 2.30 & 6.90 \\
\texttt{swh\_Latn} & 4.54 & 4.54 & 4.79 & 2.30 & 7.09 \\
\texttt{ind\_Latn} & 4.49 & 4.49 & 4.71 & 2.31 & 7.01 \\
\texttt{yor\_Latn} & 3.38 & 3.39 & 3.60 & 2.30 & 5.90 \\
\texttt{hau\_Latn} & 4.47 & 4.47 & 4.69 & 2.31 & 6.99 \\
\addlinespace[2pt]
\texttt{rus\_Cyrl} & 3.16 & 3.16 & 3.79 & 2.31 & 6.10 \\
\texttt{ukr\_Cyrl} & 3.22 & 3.22 & 3.85 & 2.30 & 6.15 \\
\texttt{srp\_Cyrl} & 3.41 & 3.41 & 3.99 & 2.30 & 6.30 \\
\texttt{kaz\_Cyrl} & 3.11 & 3.11 & 3.56 & 2.30 & 5.87 \\
\addlinespace[2pt]
\texttt{ell\_Grek} & 3.58 & 3.58 & 4.07 & 2.31 & 6.38 \\
\addlinespace[2pt]
\texttt{hye\_Armn} & 1.99 & 1.99 & 2.23 & 2.31 & 4.54 \\
\addlinespace[2pt]
\texttt{kat\_Geor} & 0.69 & 0.69 & 0.95 & 2.31 & 3.25 \\
\addlinespace[2pt]
\texttt{heb\_Hebr} & 2.42 & 2.42 & 2.64 & 2.30 & 4.94 \\
\addlinespace[2pt]
\texttt{arb\_Arab} & 3.81 & 3.83 & 3.94 & 1.57 & 5.51 \\
\texttt{pes\_Arab} & 3.91 & 3.92 & 4.06 & 1.57 & 5.63 \\
\texttt{urd\_Arab} & 4.10 & 4.11 & 4.23 & 1.57 & 5.80 \\
\addlinespace[2pt]
\texttt{cmn\_Hans} & 0.85 & 0.85 & 0.90 & 2.27 & 3.16 \\
\addlinespace[2pt]
\texttt{cmn\_Hant} & 0.75 & 0.75 & 0.78 & 2.26 & 3.04 \\
\addlinespace[2pt]
\texttt{jpn\_Jpan} & 0.62 & 0.65 & 0.67 & 2.28 & 2.95 \\
\addlinespace[2pt]
\texttt{kor\_Hang} & 1.31 & 1.32 & 1.38 & 2.29 & 3.67 \\
\addlinespace[2pt]
\texttt{hin\_Deva} & 1.01 & 1.01 & 1.47 & 1.57 & 3.05 \\
\texttt{mar\_Deva} & 0.75 & 0.75 & 1.19 & 1.57 & 2.76 \\
\addlinespace[2pt]
\texttt{ben\_Beng} & 0.83 & 0.83 & 1.53 & 1.57 & 3.10 \\
\addlinespace[2pt]
\texttt{tam\_Taml} & 1.10 & 1.10 & 1.66 & 1.57 & 3.24 \\
\addlinespace[2pt]
\texttt{tel\_Telu} & 0.90 & 0.90 & 1.37 & 1.57 & 2.94 \\
\addlinespace[2pt]
\texttt{guj\_Gujr} & 0.94 & 0.94 & 1.18 & 1.57 & 2.75 \\
\addlinespace[2pt]
\texttt{pan\_Guru} & 1.00 & 1.00 & 1.46 & 1.57 & 3.03 \\
\addlinespace[2pt]
\texttt{kan\_Knda} & 0.70 & 0.70 & 1.17 & 1.57 & 2.74 \\
\addlinespace[2pt]
\texttt{mal\_Mlym} & 0.94 & 0.94 & 1.16 & 1.57 & 2.74 \\
\addlinespace[2pt]
\texttt{tha\_Thai} & 0.48 & 0.48 & 0.76 & 2.30 & 3.07 \\
\addlinespace[2pt]
\texttt{lao\_Laoo} & 0.50 & 0.50 & 0.74 & 2.30 & 3.05 \\
\addlinespace[2pt]
\texttt{mya\_Mymr} & 0.53 & 0.53 & 0.83 & 1.58 & 2.40 \\
\addlinespace[2pt]
\texttt{khm\_Khmr} & 0.45 & 0.45 & 0.57 & 1.57 & 2.14 \\
\addlinespace[2pt]
\texttt{amh\_Ethi} & 1.00 & 1.00 & 1.07 & 2.29 & 3.37 \\
\bottomrule
\end{tabular}
\caption{Undisturbed capacity in bits per character (real-text view, no attack). $\kappa=\kappa_{\mathrm{sub}}+\kappa_{\mathrm{ins}}$ at $\pi_{\mathrm{none}}$, the identity column of Figure~\ref{fig:effective}.}
\label{tab:capacity-full}
\end{table}

\section{Substitution sets and capacity by language}\label{app:framework}

Over the frozen snapshot, the substitution sets contain 2,245 skeleton classes and 36,066 cross-script, 2,645 compatibility, 600 whitespace and 28,145 residual substitution pairs. The 1,207 code points that are confusable only with multi-character sequences cannot be swapped one for one and are excluded.

Besides the real-text capacity of \S\ref{sec:framework}, we report an alphabet view: the plain mean of $\log_2|S_\pi(c)|$ over a language's native exemplar characters. It asks how rich the writing system is rather than how rich the running text is. The two views differ: English carries 4.78 bits per character of substitution capacity in real text but 2.09 over its alphabet, because frequent characters such as the space and the common vowels have many substitutes.

Table~\ref{tab:capacity-full} gives the numbers behind \S\ref{sec:capacity}: substitution capacity under each admissibility policy of \S\ref{sec:framework}, insertion capacity, and their total. The policies nest, so $\kappa_{\mathrm{sub}}$ never decreases from left to right. Insertion capacity does not depend on the policy; it is about 2.3 bits per character where all four invisible characters are safe and about 1.57 where the two joiners are dropped.

\textbf{Stability.} The bootstrap 95\% interval on English total capacity is $[7.082, 7.089]$. Recomputing every language on the FLORES+ dev split (997 sentences) moves substitution capacity by 0.012 bits per character on average and by at most 0.060, for Yoruba.

\textbf{Characters outside the native set.} Characters outside each language's native character set carry 65\% of the substitution capacity of Simplified Chinese, 55\% of Traditional Chinese, 49\% of Japanese and all of Korean's, against 18\% for English. Foreign letters alone, such as Latin words inside Chinese text, carry 27\% of the substitution capacity of Simplified Chinese, 7\% of Japanese and 10\% of Korean, against none for English.

\textbf{Cross-script twins.} Cross-script letter twins, such as a Cyrillic a in English text, form a channel worth 1.99 bits per character on its own for English and 2.12 for Russian. Removing them under $\pi_{\mathrm{intra}}$ lowers substitution capacity only from 4.78 to 4.50 bits per character for English and from 3.79 to 3.16 for Russian (Table~\ref{tab:capacity-full}), because per-family capacities do not add (\S\ref{sec:framework}): English letters keep their compatibility variants, which $\pi_{\mathrm{intra}}$ admits.

\begin{table*}[ht!]
\centering\footnotesize
\setlength{\tabcolsep}{4pt}
\begin{tabular}{@{}lcccccccccccc@{}}
\toprule
 & \multicolumn{6}{c}{Substitution track, $s_{\mathrm{sub}}$} & \multicolumn{6}{c}{Both tracks, $s_{\mathrm{total}}$} \\
\cmidrule(lr){2-7}\cmidrule(l){8-13}
Code & NFC & C-P & NFKC & MS-HR & MS-SS & Skel. & NFC & C-P & NFKC & MS-HR & MS-SS & Skel. \\
\midrule
\texttt{eng\_Latn} & 1.00 & 1.00 & 0.59 & 0.94 & 0.94 & 0.47 & 1.00 & 0.67 & 0.72 & 0.96 & 0.96 & 0.64 \\
\texttt{spa\_Latn} & 1.00 & 1.00 & 0.61 & 0.94 & 0.94 & 0.47 & 1.00 & 0.67 & 0.74 & 0.96 & 0.96 & 0.64 \\
\texttt{fra\_Latn} & 1.00 & 1.00 & 0.60 & 0.95 & 0.95 & 0.47 & 1.00 & 0.66 & 0.73 & 0.96 & 0.96 & 0.65 \\
\texttt{deu\_Latn} & 1.00 & 1.00 & 0.58 & 0.95 & 0.95 & 0.47 & 1.00 & 0.66 & 0.72 & 0.97 & 0.97 & 0.65 \\
\texttt{por\_Latn} & 1.00 & 1.00 & 0.61 & 0.94 & 0.94 & 0.47 & 1.00 & 0.67 & 0.74 & 0.96 & 0.96 & 0.64 \\
\texttt{vie\_Latn} & 0.99 & 0.99 & 0.58 & 0.96 & 0.96 & 0.49 & 1.00 & 0.61 & 0.74 & 0.97 & 0.97 & 0.69 \\
\texttt{tur\_Latn} & 1.00 & 1.00 & 0.59 & 0.95 & 0.95 & 0.44 & 1.00 & 0.66 & 0.73 & 0.96 & 0.96 & 0.63 \\
\texttt{swh\_Latn} & 1.00 & 1.00 & 0.58 & 0.95 & 0.95 & 0.47 & 1.00 & 0.67 & 0.71 & 0.97 & 0.97 & 0.64 \\
\texttt{ind\_Latn} & 1.00 & 1.00 & 0.55 & 0.96 & 0.96 & 0.48 & 1.00 & 0.67 & 0.70 & 0.97 & 0.97 & 0.65 \\
\texttt{yor\_Latn} & 0.99 & 0.99 & 0.61 & 0.94 & 0.94 & 0.48 & 0.99 & 0.60 & 0.76 & 0.96 & 0.96 & 0.68 \\
\texttt{hau\_Latn} & 0.99 & 0.99 & 0.57 & 0.96 & 0.96 & 0.49 & 1.00 & 0.67 & 0.71 & 0.97 & 0.97 & 0.66 \\
\addlinespace[2pt]
\texttt{rus\_Cyrl} & 1.00 & 1.00 & 0.77 & 0.84 & 0.84 & 0.42 & 1.00 & 0.62 & 0.86 & 0.90 & 0.90 & 0.64 \\
\texttt{ukr\_Cyrl} & 1.00 & 1.00 & 0.77 & 0.84 & 0.84 & 0.41 & 1.00 & 0.62 & 0.86 & 0.90 & 0.90 & 0.63 \\
\texttt{srp\_Cyrl} & 1.00 & 1.00 & 0.75 & 0.86 & 0.86 & 0.41 & 1.00 & 0.63 & 0.84 & 0.91 & 0.91 & 0.63 \\
\texttt{kaz\_Cyrl} & 1.00 & 1.00 & 0.75 & 0.87 & 0.87 & 0.43 & 1.00 & 0.60 & 0.85 & 0.92 & 0.92 & 0.65 \\
\addlinespace[2pt]
\texttt{ell\_Grek} & 1.00 & 1.00 & 0.68 & 0.88 & 0.88 & 0.16 & 1.00 & 0.64 & 0.80 & 0.92 & 0.92 & 0.46 \\
\addlinespace[2pt]
\texttt{hye\_Armn} & 0.99 & 0.99 & 0.68 & 0.89 & 0.89 & 0.19 & 1.00 & 0.49 & 0.84 & 0.95 & 0.95 & 0.60 \\
\addlinespace[2pt]
\texttt{kat\_Geor} & 0.99 & 0.99 & 0.77 & 0.73 & 0.73 & 0.47 & 1.00 & 0.29 & 0.93 & 0.92 & 0.92 & 0.84 \\
\addlinespace[2pt]
\texttt{heb\_Hebr} & 0.99 & 0.99 & 0.68 & 0.91 & 0.91 & 0.21 & 0.99 & 0.53 & 0.83 & 0.95 & 0.95 & 0.58 \\
\addlinespace[2pt]
\texttt{arb\_Arab} & 0.99 & 0.99 & 0.47 & 0.97 & 0.97 & 0.13 & 1.00 & 0.71 & 0.62 & 0.98 & 0.98 & 0.38 \\
\texttt{pes\_Arab} & 0.99 & 0.99 & 0.51 & 0.97 & 0.97 & 0.15 & 1.00 & 0.72 & 0.64 & 0.98 & 0.98 & 0.39 \\
\texttt{urd\_Arab} & 0.99 & 0.99 & 0.53 & 0.97 & 0.97 & 0.15 & 1.00 & 0.72 & 0.66 & 0.98 & 0.98 & 0.38 \\
\addlinespace[2pt]
\texttt{cmn\_Hans} & 0.89 & 0.89 & 0.49 & 0.97 & 0.84 & 0.46 & 0.97 & 0.25 & 0.86 & 0.99 & 0.96 & 0.85 \\
\addlinespace[2pt]
\texttt{cmn\_Hant} & 0.83 & 0.83 & 0.41 & 0.98 & 0.93 & 0.46 & 0.96 & 0.21 & 0.85 & 1.00 & 0.98 & 0.86 \\
\addlinespace[2pt]
\texttt{jpn\_Jpan} & 0.92 & 0.92 & 0.36 & 0.97 & 0.78 & 0.60 & 0.98 & 0.21 & 0.86 & 0.99 & 0.95 & 0.91 \\
\addlinespace[2pt]
\texttt{kor\_Hang} & 0.98 & 0.98 & 0.72 & 0.96 & 0.89 & 0.59 & 0.99 & 0.37 & 0.90 & 0.99 & 0.96 & 0.85 \\
\addlinespace[2pt]
\texttt{hin\_Deva} & 0.98 & 0.98 & 0.82 & 0.66 & 0.66 & 0.42 & 0.99 & 0.48 & 0.91 & 0.84 & 0.84 & 0.72 \\
\texttt{mar\_Deva} & 0.99 & 0.99 & 0.83 & 0.61 & 0.61 & 0.38 & 0.99 & 0.43 & 0.93 & 0.83 & 0.83 & 0.73 \\
\addlinespace[2pt]
\texttt{ben\_Beng} & 0.99 & 0.99 & 0.87 & 0.53 & 0.53 & 0.30 & 0.99 & 0.49 & 0.93 & 0.77 & 0.77 & 0.66 \\
\addlinespace[2pt]
\texttt{tam\_Taml} & 0.99 & 0.99 & 0.91 & 0.53 & 0.53 & 0.22 & 1.00 & 0.51 & 0.95 & 0.76 & 0.76 & 0.60 \\
\addlinespace[2pt]
\texttt{tel\_Telu} & 0.99 & 0.99 & 0.82 & 0.66 & 0.66 & 0.32 & 0.99 & 0.46 & 0.92 & 0.84 & 0.84 & 0.69 \\
\addlinespace[2pt]
\texttt{guj\_Gujr} & 0.98 & 0.98 & 0.80 & 0.77 & 0.77 & 0.45 & 0.99 & 0.42 & 0.92 & 0.90 & 0.90 & 0.76 \\
\addlinespace[2pt]
\texttt{pan\_Guru} & 0.98 & 0.98 & 0.82 & 0.66 & 0.66 & 0.41 & 0.99 & 0.47 & 0.91 & 0.84 & 0.84 & 0.72 \\
\addlinespace[2pt]
\texttt{kan\_Knda} & 0.99 & 0.99 & 0.84 & 0.61 & 0.61 & 0.33 & 1.00 & 0.42 & 0.93 & 0.83 & 0.83 & 0.72 \\
\addlinespace[2pt]
\texttt{mal\_Mlym} & 0.99 & 0.99 & 0.83 & 0.81 & 0.81 & 0.28 & 1.00 & 0.42 & 0.93 & 0.92 & 0.92 & 0.70 \\
\addlinespace[2pt]
\texttt{tha\_Thai} & 1.00 & 1.00 & 0.90 & 0.64 & 0.63 & 0.18 & 1.00 & 0.25 & 0.98 & 0.91 & 0.91 & 0.79 \\
\addlinespace[2pt]
\texttt{lao\_Laoo} & 0.99 & 0.99 & 0.77 & 0.66 & 0.66 & 0.37 & 1.00 & 0.24 & 0.94 & 0.92 & 0.92 & 0.85 \\
\addlinespace[2pt]
\texttt{mya\_Mymr} & 0.99 & 0.99 & 0.81 & 0.61 & 0.61 & 0.24 & 1.00 & 0.34 & 0.93 & 0.87 & 0.87 & 0.74 \\
\addlinespace[2pt]
\texttt{khm\_Khmr} & 0.99 & 0.99 & 0.78 & 0.73 & 0.73 & 0.35 & 1.00 & 0.26 & 0.94 & 0.93 & 0.93 & 0.83 \\
\addlinespace[2pt]
\texttt{amh\_Ethi} & 0.98 & 0.98 & 0.74 & 0.95 & 0.95 & 0.56 & 0.99 & 0.31 & 0.92 & 0.98 & 0.98 & 0.86 \\
\bottomrule
\end{tabular}
\caption{Survival $s=\kappa^{\mathrm{eff}}/\kappa$ by language and attack, policy $\pi_{\mathrm{none}}$. C-P: copy-paste re-encoding; MS-HR and MS-SS: the mixed-script detector at the Highly Restrictive and Single Script levels; Skel.: skeleton mapping, which keeps invisible characters (Table~\ref{tab:ladder}). Insertion survival is not tabulated because it is constant: 1.00 for every language under every attack shown except copy-paste, where it is 0.00 for every language.}
\label{tab:rho}
\end{table*}

\section{Survival under attack}\label{app:robustness}
Table~\ref{tab:rho} gives the survival $s=\kappa^{\mathrm{eff}}/\kappa$ of \S\ref{sec:robustness} under each attack of Figure~\ref{fig:effective}, for the substitution track and for the two tracks combined. The insertion track needs no column: it survives each of them intact except copy-paste, which removes it for all 40 languages. The table also carries the stricter Single Script level of the mixed-script detector, which the main text does not discuss; its substitution survival differs from the Highly Restrictive level by more than 0.01 only for Chinese, Japanese and Korean.

Each transformation removes a different kind of substitute, so each hits a different group hardest. In substitution survival, NFKC hits Chinese, Japanese and the Arabic-script languages hardest (0.36 to 0.53), mixed-script detection the Brahmic and South-East Asian scripts (0.53 to 0.81), and skeleton mapping the Arabic-script languages, Greek, Thai, Armenian and Hebrew (0.13 to 0.21), while the Latin-script languages keep 0.44 to 0.49 under skeleton mapping.

Mixed-script detection permits Latin combined with Han, kana or Hangul at the Highly Restrictive level, so the characters outside the native set that carry much of CJK capacity (\S\ref{sec:capacity}) pass untouched. Under $\pi_{\mathrm{intra}}$, which admits no other-script substitutes, the detector finds almost nothing to remove (mean survival 0.98, against 0.84 under $\pi_{\mathrm{none}}$); the exception is Tamil (0.79), for the reason given in Appendix~\ref{app:stealth}.

Skeleton mapping splits the languages by their native script. Twenty of the 40 lose all substitution capacity from their native characters: 15 write in scripts with no compatibility variants, and in Arabic, Persian, Urdu, Hebrew and Korean the compatibility variants are themselves listed as confusables and fold too. The other 20 keep the compatibility and whitespace variants that the mapping does not merge. Every language still keeps some capacity from embedded Latin text, and all of its insertion track. The full skeleton, which also deletes default-ignorables (\S\ref{sec:framework}), removes that track too: survival then averages 0.19, below every other single transformation in all 40 languages, and the range between languages widens to 17.0$\times$, with Thai lowest.

\begin{figure*}[t!]
\centering
\includegraphics[width=0.85\textwidth]{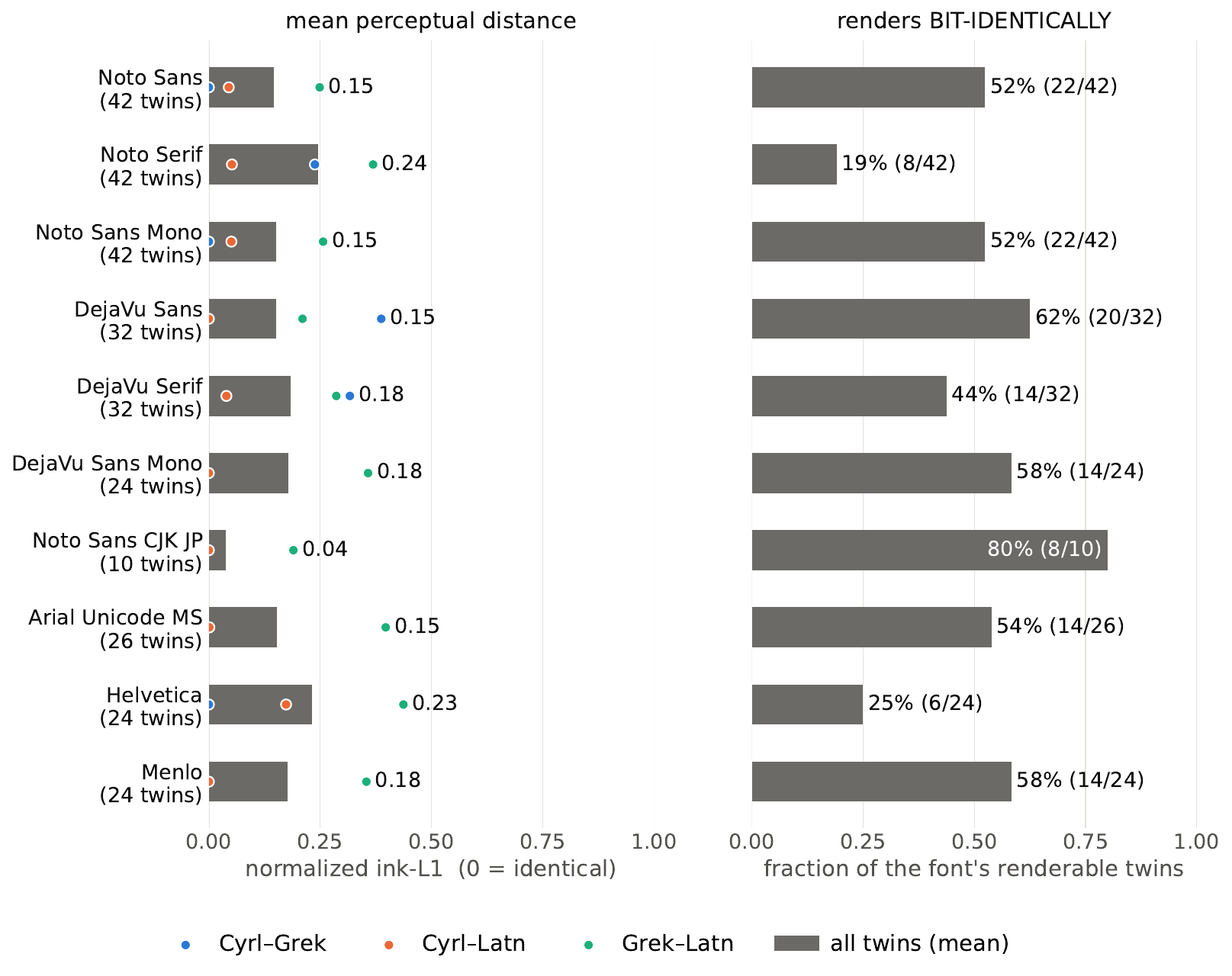}
\caption{Perceptual safety is font-relative. A genuine cross-script letter twin is a pair with the same Unicode letter name and case in two of Latin, Cyrillic and Greek; there are 21, each counted in both directions (42 in all), and each row is a panel font that contains at least one. Left: mean normalized ink difference between the two glyphs of a twin, over all twins (bar) and by script pair (dots). Right: the fraction of the font's renderable twins that render bit-identically. Glyphs are rendered in isolation, in the same font, at 48 pixels.}
\label{fig:fontdiv}
\end{figure*}

\section{Stealth}\label{app:stealth}
This appendix gives the stealth analysis summarized in \S\ref{sec:robustness}. Capacity counts every substitute, but a substitute only helps if no one notices it. We check three ways it can be noticed: a mixed-script detector, which reads script labels and never a font; Unicode's identifier status, which marks some characters as Restricted~\citep{uts39}; and the reader's eye, which sees glyphs that depend on the installed fonts. We keep the checks apart because they disagree. A cross-script twin (family~i) can look identical to the original yet trip the detector, since it carries a foreign script label; a compatibility variant (family~ii) never trips the detector, but is often drawn wider or styled. A language can look stealthy under one check and loud under another, and its rank among the 40 changes with the check, so no single stealth score can describe it. Table~\ref{tab:stealth} reports all three checks per language; its eye-invisible share is given both at a 0.02 threshold and for bit-identical substitutes, so that no conclusion rests on the threshold.

Against the detector, the embedder's choices matter most. Substituting at every possible position under $\pi_{\mathrm{none}}$ gets 94.7\% of English tokens flagged at the Highly Restrictive level, against none for unmarked text, because almost every word ends up mixing scripts; skipping substitutes that would break a token into two scripts brings this down to 0.5\%. The strict policy alone is not enough. Under $\pi_{\mathrm{intra}}$, substituting everywhere still gets 0.7\% to 80.1\% of tokens flagged, worst for Tamil: the virama sign, 14.4\% of Tamil characters, has an Inherited-script homoglyph that $\pi_{\mathrm{intra}}$ admits but whose Script\_Extensions do not list Tamil, so any word containing it stops being single-script. Scripts that lean on combining marks are exposed in this way through Unicode's script assignments, not through anything about the languages. The exposure is avoidable: skipping token-breaking substitutes keeps every language within 0.001 of unmarked text under $\pi_{\mathrm{intra}}$.

Identifier status cuts the other way. Most of English's substitution capacity already sits in substitutes that UTS~\#39 marks as Restricted (\S\ref{sec:capacity}), and the stealthier policies make this worse: dropping cross-script substitutes removes the Allowed foreign letters and keeps the Restricted compatibility variants. An embedder that avoids the mixed-script detector therefore puts more of its bits into exactly the characters an identifier filter would flag.

To a reader, homoglyphs are less identical than their name suggests. We render each character and its substitute and measure how much their ink differs, from 0 for identical outlines; whitespace, which has no ink, is compared by width. On a stack that models a typical reader, Noto fonts for the scripts UAX~\#31 recommends~\citep{uax31} plus CJK, the mean difference is between 0.60 and 0.65 for every family, because many substitutes are drawn in another typeface or style. Picking the closest substitute for each character helps a great deal, cutting English from 0.62 to 0.14 and Arabic from 0.52 to 0.03. These distances compare glyphs drawn in isolation, which flatters Arabic: its nearest substitutes are mostly isolated presentation forms, which match a letter drawn alone but not one joined inside a word. Eye-invisible substitutes, within 0.02 of the original, are rare and uneven: on a stack with full Noto coverage they carry 13.2\% of the cross-script channel for English and 2.8\% for Russian, but 0.2\% for Bengali and Arabic, because fonts reuse one drawing for two characters mainly across Latin, Cyrillic and Greek. Even this depends on the font: among the three fonts in Figure~\ref{fig:fontdiv} that render every twin, between 19\% and 52\% are drawn identically, so ``visually identical'' only means something for a named font.

A substitute the reader's fonts cannot draw appears as an empty box, so it is not usable capacity either. On the typical-reader stack, coverage keeps up to 0.94 of substitution capacity for the best-covered languages, but 0.75 for English and only 0.58 for Bengali. What is lost differs by script. Bengali loses mainly one sibling script, Tirhuta, which the stack lacks because UAX~\#31 does not recommend it; Tirhuta alone accounts for 0.57 bits per character, 89\% of Bengali's loss. English loses mainly styled symbols from the Common script, 92\% of its loss.

The insertion track faces a similar test. Beyond the joiners already excluded (\S\ref{sec:framework}), whether an invisible character stays invisible depends on the software that shapes the text. ZERO WIDTH SPACE and WORD JOINER leave an Arabic word joined under OpenType shaping and in Apple's own CoreText engine, but break the join in every Apple AAT font we tested when the fonts are shaped by HarfBuzz, which our renderer (Pillow with libraqm) uses. Arabic insertion capacity is therefore 1.57 bits per character under OpenType and CoreText, and 0 under HarfBuzz's AAT shaping. As with capacity, the cost of staying hidden follows script: Tamil pays through Unicode's script assignments, Bengali through a sibling script that readers' fonts lack, and Arabic through how one shaping engine treats invisible characters.

\begin{table*}[ht!]
\centering\footnotesize
\setlength{\tabcolsep}{5pt}
\begin{tabular}{@{}lccccccccc@{}}
\toprule
 & \multicolumn{3}{c}{Detector: tokens flagged (\%)} & Identifier & \multicolumn{2}{c}{Distance} & Coverage & \multicolumn{2}{c}{Eye-invisible share} \\
\cmidrule(lr){2-4}\cmidrule(lr){5-5}\cmidrule(lr){6-7}\cmidrule(lr){8-8}\cmidrule(l){9-10}
Code & loud & quiet & loud, $\pi_{\mathrm{intra}}$ & restricted & typical & nearest & rendered & $\le 0.02$ & identical \\
\midrule
\texttt{eng\_Latn} & 94.7 & 0.5 & 2.2 & 0.70 & 0.62 & 0.14 & 0.75 & 0.16 & 0.15 \\
\texttt{spa\_Latn} & 79.3 & 0.4 & 0.7 & 0.69 & 0.61 & 0.11 & 0.75 & 0.16 & 0.15 \\
\texttt{fra\_Latn} & 81.4 & 0.4 & 6.0 & 0.71 & 0.61 & 0.13 & 0.75 & 0.15 & 0.14 \\
\texttt{deu\_Latn} & 94.4 & 0.6 & 2.3 & 0.72 & 0.62 & 0.14 & 0.73 & 0.15 & 0.14 \\
\texttt{por\_Latn} & 82.0 & 0.5 & 1.4 & 0.70 & 0.60 & 0.11 & 0.75 & 0.17 & 0.15 \\
\texttt{vie\_Latn} & 93.6 & 0.5 & 1.2 & 0.76 & 0.58 & 0.14 & 0.75 & 0.22 & 0.21 \\
\texttt{tur\_Latn} & 97.0 & 0.5 & 4.5 & 0.70 & 0.62 & 0.15 & 0.74 & 0.14 & 0.12 \\
\texttt{swh\_Latn} & 90.2 & 0.6 & 1.5 & 0.71 & 0.63 & 0.14 & 0.73 & 0.16 & 0.14 \\
\texttt{ind\_Latn} & 98.4 & 0.7 & 2.1 & 0.73 & 0.63 & 0.16 & 0.72 & 0.14 & 0.12 \\
\texttt{yor\_Latn} & 87.2 & 0.3 & 11.9 & 0.74 & 0.62 & 0.20 & 0.76 & 0.19 & 0.17 \\
\texttt{hau\_Latn} & 92.1 & 0.5 & 2.7 & 0.74 & 0.62 & 0.15 & 0.73 & 0.15 & 0.14 \\
\addlinespace[2pt]
\texttt{rus\_Cyrl} & 80.5 & 0.5 & 2.9 & 0.72 & 0.42 & 0.12 & 0.68 & 0.08 & 0.07 \\
\texttt{ukr\_Cyrl} & 80.1 & 0.6 & 3.1 & 0.72 & 0.43 & 0.12 & 0.68 & 0.11 & 0.07 \\
\texttt{srp\_Cyrl} & 71.8 & 0.8 & 1.7 & 0.72 & 0.42 & 0.10 & 0.68 & 0.10 & 0.09 \\
\texttt{kaz\_Cyrl} & 90.0 & 0.5 & 2.4 & 0.74 & 0.43 & 0.12 & 0.64 & 0.11 & 0.08 \\
\addlinespace[2pt]
\texttt{ell\_Grek} & 83.2 & 0.6 & 1.0 & 0.67 & 0.44 & 0.12 & 0.69 & 0.17 & 0.15 \\
\addlinespace[2pt]
\texttt{hye\_Armn} & 84.7 & 0.5 & 9.0 & 0.73 & 0.49 & 0.21 & 0.69 & 0.11 & 0.11 \\
\addlinespace[2pt]
\texttt{kat\_Geor} & 74.3 & 0.8 & 2.9 & 0.71 & 0.61 & 0.40 & 0.92 & 0.25 & 0.24 \\
\addlinespace[2pt]
\texttt{heb\_Hebr} & 76.3 & 0.6 & 46.3 & 0.71 & 0.63 & 0.44 & 0.88 & 0.12 & 0.12 \\
\addlinespace[2pt]
\texttt{arb\_Arab} & 59.5 & 0.5 & 5.8 & 0.78 & 0.52 & 0.03 & 0.79 & 0.29 & 0.29 \\
\texttt{pes\_Arab} & 64.6 & 0.4 & 1.3 & 0.76 & 0.51 & 0.02 & 0.80 & 0.30 & 0.30 \\
\texttt{urd\_Arab} & 56.1 & 0.3 & 5.9 & 0.73 & 0.53 & 0.02 & 0.83 & 0.31 & 0.31 \\
\addlinespace[2pt]
\texttt{cmn\_Hans} & 36.4 & 5.8 & 18.6 & 0.83 & 0.42 & 0.18 & 0.84 & 0.26 & 0.25 \\
\addlinespace[2pt]
\texttt{cmn\_Hant} & 25.0 & 4.4 & 9.9 & 0.86 & 0.38 & 0.19 & 0.85 & 0.29 & 0.28 \\
\addlinespace[2pt]
\texttt{jpn\_Jpan} & 63.7 & 0.3 & 13.9 & 0.85 & 0.64 & 0.46 & 0.90 & 0.12 & 0.11 \\
\addlinespace[2pt]
\texttt{kor\_Hang} & 18.0 & 0.1 & 3.7 & 0.92 & 0.50 & 0.12 & 0.88 & 0.33 & 0.33 \\
\addlinespace[2pt]
\texttt{hin\_Deva} & 76.2 & 0.4 & 20.2 & 0.72 & 0.41 & 0.18 & 0.90 & 0.26 & 0.26 \\
\texttt{mar\_Deva} & 87.6 & 0.3 & 17.4 & 0.66 & 0.40 & 0.21 & 0.90 & 0.23 & 0.23 \\
\addlinespace[2pt]
\texttt{ben\_Beng} & 81.2 & 0.4 & 14.3 & 0.87 & 0.38 & 0.16 & 0.58 & 0.19 & 0.19 \\
\addlinespace[2pt]
\texttt{tam\_Taml} & 93.9 & 0.5 & 80.1 & 0.57 & 0.47 & 0.38 & 0.94 & 0.19 & 0.18 \\
\addlinespace[2pt]
\texttt{tel\_Telu} & 92.9 & 0.7 & 3.1 & 0.56 & 0.58 & 0.45 & 0.91 & 0.19 & 0.18 \\
\addlinespace[2pt]
\texttt{guj\_Gujr} & 62.3 & 0.4 & 16.8 & 0.74 & 0.47 & 0.19 & 0.91 & 0.27 & 0.27 \\
\addlinespace[2pt]
\texttt{pan\_Guru} & 82.1 & 0.4 & 18.8 & 0.72 & 0.44 & 0.23 & 0.90 & 0.26 & 0.26 \\
\addlinespace[2pt]
\texttt{kan\_Knda} & 93.7 & 0.7 & 2.3 & 0.54 & 0.57 & 0.46 & 0.93 & 0.20 & 0.20 \\
\addlinespace[2pt]
\texttt{mal\_Mlym} & 88.9 & 0.6 & 3.2 & 0.56 & 0.56 & 0.36 & 0.90 & 0.16 & 0.16 \\
\addlinespace[2pt]
\texttt{tha\_Thai} & 87.2 & 2.7 & 5.5 & 0.27 & 0.51 & 0.46 & 0.94 & 0.08 & 0.08 \\
\addlinespace[2pt]
\texttt{lao\_Laoo} & 79.8 & 2.1 & 20.1 & 0.56 & 0.65 & 0.47 & 0.86 & 0.15 & 0.14 \\
\addlinespace[2pt]
\texttt{mya\_Mymr} & 88.0 & 0.8 & 16.3 & 0.56 & 0.57 & 0.38 & 0.86 & 0.15 & 0.15 \\
\addlinespace[2pt]
\texttt{khm\_Khmr} & 84.8 & 2.5 & 39.5 & 0.59 & 0.63 & 0.45 & 0.87 & 0.17 & 0.17 \\
\addlinespace[2pt]
\texttt{amh\_Ethi} & 23.2 & 0.5 & 2.3 & 0.92 & 0.47 & 0.14 & 0.88 & 0.36 & 0.36 \\
\bottomrule
\end{tabular}
\caption{Stealth per language, $\pi_{\mathrm{none}}$ unless marked. Detector: \% of tokens flagged at the Highly Restrictive level, substituting everywhere (loud) or skipping token-breaking substitutes (quiet). Identifier: share of substitution capacity in \emph{Restricted} substitutes. On the typical-reader font stack: Distance, mean normalized ink difference to a typical and the nearest substitute (0 = identical); Coverage, share of substitution capacity whose substitutes render; Eye-invisible, share in substitutes within distance 0.02 of the original, and in bit-identical ones. Both eye-invisible shares come mostly from substitutes for the space character, and the check counts CARRIAGE RETURN, a line break in plain text, as identical to a space.}\label{tab:stealth}
\end{table*}

\section{Keyed Instrument and Detection Length}\label{app:keyed}

The embedder walks the text, visiting at each position $i$ a substitution slot with alphabet $S_\pi(c_i)$ and then the insertion gap after $i$ with alphabet $M_L\cup\{\varnothing\}$; index 0 is the identity option on both tracks. A PRNG seeded by the key (HMAC-SHA256 with rejection sampling, so that every symbol is exactly equiprobable) and labelled by the language, policy, track, position and the five preceding cover characters draws one symbol per slot. Content-keying the label matters: a position-only label makes the schedule a function of the key alone, so every document draws the same symbols and the null variance collapses; only false-positive calibration exposed this. Because the per-slot alphabets are exactly the admissible sets of \S\ref{sec:framework}, summing $\log_2 q$ over the instrument's own slot walk reproduces $\kappa$ and $\kappa^{\mathrm{eff}}$ to within $10^{-14}$. The detector is cover-informed by necessity rather than convenience: substitutes carry their own admissible sets, so a blind walk mis-derives the alphabet at up to 85\% of substitutable positions in homoglyph-rich scripts.

Under the null of ``not watermarked with this key'' each slot matches with probability $p_i=1/q_i$; under an attack $\phi$, $p_i$ is the share of the alphabet that $\phi$ maps onto the identity's image, so a destroyed slot has $p_i=1$ and contributes no evidence. We declare a watermark when the $z$-score of the match count $X$, $z = (X - \sum_i p_i)/\sqrt{\sum_i p_i(1-p_i)}$, exceeds the 99th percentile of null $z$-scores on unwatermarked text, set separately for each language, condition and span length from 400 to 1,000 trials. We use this empirical threshold rather than a normal quantile because the normal approximation is anti-conservative at short lengths: a normal 1\% threshold gives a false-positive rate of 0.054 for Arabic at two code points on clean text. On fresh keys the thresholds give a mean false-positive rate of 1.01\% (0.96\% to 1.04\% across conditions), and 0.98\% to 1.32\% across conditions at the span length nearest each language's $n_{\min}$. Single cells scatter around this, up to 5.0\% for Indonesian at 128 code points under mixed-script detection, because each threshold is set, and each rate measured, on 400 to 1,000 trials. 

The detector rebuilds its schedule from the received document, and an attack changes the characters the labels are computed from. Keying labels on the raw received characters is self-defeating: 13.9\% of labels survive attack, the detection statistic drops by 19.8 standard units, and the required length is censored beyond 256 code points for 25 of the 40 languages. Keying on the skeleton keeps 59.7\% of labels at a cost of 9.9 units. Keying on the canonical fold keeps 98.5\% at a cost of 0.36, and we use this canonical keying throughout. The fold decomposes each character with the frozen Unicode 17.0.0 compatibility decompositions, maps it to its UTS~\#39 skeleton prototype and turns every White\_Space character into U+0020; default-ignorable characters are deleted, which removes all four insertion characters.

Spans are sampled by equal-width strata across the document, and spans with no usable slot are scored at zero evidence rather than skipped, since skipping them would condition on the text being watermarkable. In \S\ref{sec:keyed}, $\kappa^{\mathrm{eff}}$ is evaluated at $\pi_{\mathrm{lang}}$ and at the gap rate of a span of length $L$, $(L-1)/L$, at the measured $n_{\min}$; \S\ref{sec:capacity}-\S\ref{sec:robustness} use $\pi_{\mathrm{none}}$ and the corpus gap rate. For English the two are related exactly: $7.09\to6.80$ by the policy step and then $\to5.63$ by the gap-rate step. Each is correct for its analysis, and no table or figure mixes them. Under this convention the between-language spread of $\kappa^{\mathrm{eff}}$ is $3.28\times$ on clean text, $10.11\times$ under copy-paste and $2.52\times$ under NFKC.

On clean text the median $n_{\min}$ is 2.83 code points, English needs 1.95, and no language is censored: a $2.7\times$ span of detection length against a $3.3\times$ span of capacity. One might expect family~(ii) share (family~(ii)'s capacity as a fraction of $\kappa_{\mathrm{sub}}$) to explain the weaker NFKC correlation, but that is only half right: at $\pi_{\mathrm{lang}}$ it predicts the capacity loss strongly (Spearman $-0.834$ with substitution survival) but the ratio of measured to predicted $n_{\min}$ only moderately ($+0.437$).

The combined attack applies the canonical fold to each character; its largest mean $z$ is 0.288 and its largest AUC 0.589. Skeleton mapping alone is not a floor, because the insertion track and some compatibility and whitespace variants survive it. Two nulls exist, unwatermarked text and watermarked text read with the wrong key, and the thresholds use only the first. The two coincide on clean text, where no language has a collapsed image class, and stay close under copy-paste and mixed-script detection, but under NFKC the thresholds flag far more wrong-key text in Armenian and the four Cyrillic-script languages. Finally, 22 of the 40 languages already write with characters from the insertion alphabet, which makes an inserted invisible next to a native one unattributable; the decoder resolves runs rather than characters, and the effect is material only for Khmer, which writes word breaks with ZERO WIDTH SPACE and pays a 5.03\% discount on decodable capacity. Persian has 1,739 native non-joiners and no ambiguous gap, because the joiner rule already removed the non-joiner from its safe set.

Table~\ref{tab:nmin} gives $n_{\min}$ with 95\% bootstrap intervals, under the four attacks of Figure~\ref{fig:nmin}. Every value uses policy $\pi_{\mathrm{lang}}$ and canonical keying, the convention of \S\ref{sec:keyed}, and no cell is censored.

\section{Reproducibility}\label{app:repro}
We use Unicode 17.0.0 data files~\citep{unicodestandard, uts39}, the CLDR 48.2.0 exemplar characters~\citep{cldr}, and FLORES+ 4.5. The devtest split (1,012 sentences per language) gives every reported number and the dev split (997 sentences) the stability check of Appendix~\ref{app:framework}. The Noto and DejaVu fonts behind the perceptual measures are pinned by release and file hash; the three system fonts in Figure~\ref{fig:fontdiv} are recorded by hash but cannot be redistributed. Resampling is seeded: the capacity and detection-length bootstraps use seed 12345, the permutation tests of \S\ref{sec:fairness} use 10,000 permutations at each of the seeds 20260801 and 314159, the cluster bootstrap uses 10,000 resamples at seed 271828, and the fresh-key check of Appendix~\ref{app:keyed} uses seed 20261011.

\begin{table*}[t]
\centering\footnotesize
\setlength{\tabcolsep}{5pt}
\begin{tabular}{@{}lr@{\ \,}lr@{\ \,}lr@{\ \,}lr@{\ \,}l@{}}
\toprule
Code & \multicolumn{2}{c}{Clean} & \multicolumn{2}{c}{Copy-paste} & \multicolumn{2}{c}{Mixed-script (HR)} & \multicolumn{2}{c}{NFKC} \\
\midrule
\texttt{eng\_Latn} & 1.95 & [1.93, 1.96] & 1.96 & [1.94, 1.98] & 1.95 & [1.93, 1.96] & 4.56 & [3.94, 4.98] \\
\texttt{spa\_Latn} & 1.90 & [1.87, 1.93] & 2.05 & [1.88, 2.93] & 1.90 & [1.87, 1.93] & 3.97 & [3.94, 4.47] \\
\texttt{fra\_Latn} & 1.93 & [1.91, 1.94] & 1.94 & [1.92, 2.89] & 1.93 & [1.91, 1.94] & 3.94 & [2.94, 3.97] \\
\texttt{deu\_Latn} & 1.92 & [1.91, 1.93] & 1.93 & [1.92, 1.97] & 1.92 & [1.91, 1.93] & 2.96 & [2.94, 3.48] \\
\texttt{por\_Latn} & 1.95 & [1.88, 1.98] & 1.96 & [1.89, 2.00] & 1.95 & [1.88, 1.98] & 3.95 & [3.90, 3.97] \\
\texttt{vie\_Latn} & 1.96 & [1.90, 2.76] & 2.80 & [1.98, 2.96] & 1.96 & [1.90, 2.76] & 3.85 & [2.95, 3.95] \\
\texttt{tur\_Latn} & 1.98 & [1.94, 2.91] & 2.35 & [1.98, 3.46] & 1.98 & [1.94, 2.91] & 10.57 & [2.99, 12.82] \\
\texttt{swh\_Latn} & 1.87 & [1.83, 1.92] & 1.88 & [1.84, 1.94] & 1.87 & [1.83, 1.92] & 3.78 & [2.95, 3.90] \\
\texttt{ind\_Latn} & 1.85 & [1.00, 1.87] & 1.85 & [1.00, 1.88] & 1.85 & [1.00, 1.87] & 2.93 & [2.91, 3.93] \\
\texttt{yor\_Latn} & 2.42 & [1.96, 2.56] & 4.32 & [3.90, 4.73] & 2.56 & [2.00, 2.66] & 3.94 & [3.86, 3.96] \\
\texttt{hau\_Latn} & 1.95 & [1.93, 1.98] & 1.96 & [1.93, 2.06] & 1.95 & [1.93, 1.98] & 2.95 & [2.94, 3.95] \\
\addlinespace[2pt]
\texttt{rus\_Cyrl} & 1.87 & [1.86, 2.70] & 3.70 & [3.50, 4.66] & 1.87 & [1.86, 2.70] & 2.35 & [2.00, 2.87] \\
\texttt{ukr\_Cyrl} & 2.42 & [1.99, 2.82] & 3.05 & [2.94, 3.59] & 2.45 & [1.99, 2.82] & 2.84 & [2.71, 3.00] \\
\texttt{srp\_Cyrl} & 1.91 & [1.88, 2.10] & 2.98 & [2.88, 4.04] & 1.91 & [1.88, 2.10] & 2.61 & [2.00, 2.77] \\
\texttt{kaz\_Cyrl} & 2.94 & [2.83, 3.51] & 3.87 & [3.67, 5.76] & 2.94 & [2.83, 3.46] & 2.93 & [2.83, 3.35] \\
\addlinespace[2pt]
\texttt{ell\_Grek} & 1.90 & [1.88, 2.15] & 2.50 & [2.29, 2.90] & 1.90 & [1.88, 2.15] & 3.81 & [3.63, 3.95] \\
\addlinespace[2pt]
\texttt{hye\_Armn} & 2.73 & [2.67, 2.79] & 5.60 & [4.23, 5.93] & 2.74 & [2.68, 2.80] & 2.87 & [2.85, 3.74] \\
\addlinespace[2pt]
\texttt{kat\_Geor} & 3.93 & [2.92, 3.95] & 22.40 & [20.73, 23.88] & 3.93 & [2.92, 3.95] & 3.96 & [3.87, 4.95] \\
\addlinespace[2pt]
\texttt{heb\_Hebr} & 2.72 & [2.61, 2.88] & 4.66 & [3.87, 5.21] & 2.83 & [2.78, 3.10] & 3.92 & [3.86, 4.93] \\
\addlinespace[2pt]
\texttt{arb\_Arab} & 1.90 & [1.87, 1.94] & 1.96 & [1.91, 2.38] & 1.92 & [1.87, 1.95] & 3.86 & [3.80, 4.05] \\
\texttt{pes\_Arab} & 1.85 & [1.84, 1.87] & 1.85 & [1.84, 2.32] & 1.85 & [1.84, 1.87] & 4.67 & [3.87, 4.87] \\
\texttt{urd\_Arab} & 1.92 & [1.92, 1.93] & 2.67 & [2.42, 2.83] & 1.92 & [1.92, 1.93] & 3.70 & [3.57, 3.77] \\
\addlinespace[2pt]
\texttt{cmn\_Hans} & 3.89 & [3.74, 3.91] & 43.95 & [32.82, 45.16] & 3.89 & [3.74, 3.91] & 3.94 & [3.94, 4.69] \\
\addlinespace[2pt]
\texttt{cmn\_Hant} & 3.76 & [2.91, 3.89] & 31.65 & [28.82, 38.63] & 3.76 & [2.91, 3.90] & 3.94 & [3.94, 4.67] \\
\addlinespace[2pt]
\texttt{jpn\_Jpan} & 3.91 & [2.93, 3.94] & 36.99 & [30.95, 41.34] & 3.91 & [2.93, 3.94] & 3.95 & [3.94, 3.96] \\
\addlinespace[2pt]
\texttt{kor\_Hang} & 3.83 & [2.90, 3.88] & 7.99 & [7.89, 11.52] & 3.84 & [2.90, 3.89] & 3.96 & [3.93, 3.97] \\
\addlinespace[2pt]
\texttt{hin\_Deva} & 3.86 & [3.84, 5.66] & 11.77 & [11.26, 15.25] & 3.86 & [3.84, 4.80] & 4.92 & [4.71, 5.70] \\
\texttt{mar\_Deva} & 4.88 & [4.85, 4.91] & 15.82 & [14.83, 15.96] & 4.90 & [4.87, 5.88] & 4.97 & [4.93, 5.96] \\
\addlinespace[2pt]
\texttt{ben\_Beng} & 4.82 & [3.89, 5.67] & 15.42 & [14.61, 18.48] & 4.84 & [3.89, 5.76] & 5.71 & [4.92, 5.91] \\
\addlinespace[2pt]
\texttt{tam\_Taml} & 3.78 & [3.59, 3.97] & 11.33 & [10.67, 13.57] & 3.87 & [3.68, 4.68] & 3.86 & [3.67, 4.14] \\
\addlinespace[2pt]
\texttt{tel\_Telu} & 4.91 & [4.88, 4.95] & 12.75 & [11.87, 20.41] & 4.91 & [4.88, 4.95] & 4.95 & [4.90, 5.93] \\
\addlinespace[2pt]
\texttt{guj\_Gujr} & 4.73 & [3.83, 4.77] & 14.26 & [11.49, 15.53] & 4.76 & [3.85, 4.79] & 4.92 & [4.67, 4.93] \\
\addlinespace[2pt]
\texttt{pan\_Guru} & 3.87 & [3.86, 4.79] & 13.03 & [11.84, 13.93] & 3.88 & [3.87, 4.84] & 4.93 & [3.93, 4.94] \\
\addlinespace[2pt]
\texttt{kan\_Knda} & 4.91 & [3.91, 4.94] & 19.69 & [15.62, 23.08] & 4.91 & [3.91, 4.93] & 4.92 & [3.95, 5.88] \\
\addlinespace[2pt]
\texttt{mal\_Mlym} & 4.82 & [3.88, 4.85] & 12.37 & [11.54, 14.74] & 4.82 & [3.88, 4.85] & 4.89 & [4.60, 4.91] \\
\addlinespace[2pt]
\texttt{tha\_Thai} & 3.92 & [3.92, 3.94] & 43.01 & [37.08, 45.12] & 3.92 & [3.92, 3.94] & 3.93 & [3.92, 3.94] \\
\addlinespace[2pt]
\texttt{lao\_Laoo} & 3.94 & [3.93, 4.94] & 92.96 & [84.16, 141.66] & 3.94 & [3.93, 4.94] & 3.95 & [3.94, 4.96] \\
\addlinespace[2pt]
\texttt{mya\_Mymr} & 4.92 & [4.91, 5.89] & 29.78 & [22.74, 32.82] & 4.92 & [4.91, 5.89] & 4.92 & [4.91, 5.90] \\
\addlinespace[2pt]
\texttt{khm\_Khmr} & 4.95 & [4.92, 8.00] & 61.74 & [57.26, 80.69] & 4.96 & [4.93, 8.50] & 4.96 & [4.93, 8.50] \\
\addlinespace[2pt]
\texttt{amh\_Ethi} & 3.85 & [2.90, 3.87] & 12.74 & [11.86, 13.61] & 3.85 & [2.90, 3.87] & 3.94 & [3.90, 4.90] \\
\bottomrule
\end{tabular}
\caption{Detection length $n_{\min}$ in code points, the shortest span that reaches a true-positive rate of 95\% at a false-positive rate of 1\%, with 95\% bootstrap intervals. Policy $\pi_{\mathrm{lang}}$, canonical keying.}
\label{tab:nmin}
\end{table*}

\end{document}